\documentclass{article}

\PassOptionsToPackage{numbers, compress}{natbib}
\usepackage[final]{neurips_2025}
\usepackage{macros}
\usepackage{natbib}

\usepackage[utf8]{inputenc}
\usepackage{longtable} % For tables spanning multiple pages
\usepackage{ragged2e}  % For better alignment in paragraph columns (\RaggedRight)
\usepackage{geometry}  % To potentially adjust margins if needed
\usepackage{dirtree}

\usepackage[utf8]{inputenc} % allow utf-8 input
\usepackage[T1]{fontenc}    % use 8-bit T1 fonts
\usepackage{url}            % simple URL typesetting
\usepackage{booktabs}       % professional-quality tables
\usepackage{amsfonts}       % blackboard math symbols
\usepackage{nicefrac}       % compact symbols for 1/2, etc.
\usepackage{microtype}      % microtypography
\usepackage{xcolor}         % colors

\usepackage{xcolor}
\usepackage{listings}
\definecolor{codebg}{RGB}{245,245,245}
\usepackage{bbding}
\definecolor{green}{HTML}{009B55}
\usepackage{microtype}
\usepackage{graphicx}
\usepackage{subfigure}
\usepackage{threeparttable}
\usepackage{booktabs} % for professional tables
\usepackage{xspace}
\usepackage[ruled,vlined]{algorithm2e}
\usepackage{macros}
\usepackage{wrapfig}
\usepackage{lipsum}
\usepackage{amsfonts}       % blackboard math symbols
\usepackage{nicefrac}       % compact symbols for 1/2, etc.
\usepackage{xcolor, colortbl}         % colors
\usepackage{multicol}
\usepackage{multirow}
\usepackage{enumitem}

\usepackage{listings}
\usepackage{pythonhighlight}

\usepackage{titletoc}
\usepackage{mathrsfs}
\usepackage{upquote}
\usepackage[linewidth=0.5pt,leftmargin=1em,rightmargin=1em,innertopmargin=5pt,
    innerbottommargin=5pt,
    innerrightmargin=5pt,
    innerleftmargin=5pt]{mdframed}
\usepackage[toc,page,header]{appendix}

\usepackage{times}
\usepackage{latexsym}
\usepackage{tabularx}
\usepackage{multirow}
\usepackage{booktabs} % To thicken table lines
\usepackage{enumitem}
\usepackage{amsmath}

\usepackage{inconsolata}
\usepackage{graphicx}

\usepackage[utf8]{inputenc} % allow utf-8 input
\usepackage[T1]{fontenc}    % use 8-bit T1 fonts
\usepackage{url}            % simple URL typesetting
\usepackage{booktabs}       % professional-quality tables
\usepackage{amsfonts}       % blackboard math symbols
\usepackage{nicefrac}       % compact symbols for 1/2, etc.
\usepackage{microtype}      % microtypography
\usepackage{xcolor}         % colors
\usepackage{multicol}
\usepackage{transparent}
\usepackage{array}
\usepackage{bm}
\usepackage[colorlinks=true, citecolor=green, urlcolor=green]{hyperref}

\definecolor{nblue}{cmyk}{0.95,0.0,0.2,0.2}

\newcommand{\method}{\texttt{MLE-Dojo}\xspace}

\usepackage{xspace}
\usepackage{cleveref}
\usepackage{colortbl}
\PassOptionsToPackage{dvipsnames,x11names,table}{xcolor}
\usepackage{listings}
\definecolor{green}{HTML}{009B55}

\usepackage{listings}
\usepackage{pythonhighlight}
\usepackage{fancyvrb}
\RecustomVerbatimCommand{\VerbatimInput}{VerbatimInput}{fontsize=\footnotesize,
 frame=single,  
 framesep=0.5em, % separation between frame and text
 labelposition=topline,
}

\usepackage{tcolorbox}
\tcbuselibrary{listings}
\usepackage{alltt}
\tcbuselibrary{breakable}

\title{\method: Interactive Environments for Empowering LLM Agents in Machine Learning Engineering }

\author{%
  \textbf{Rushi Qiang}\textsuperscript{\textnormal{\dag}}\thanks{Equal contribution.}\textnormal{,} 
  \textbf{Yuchen Zhuang}\textsuperscript{\textnormal{\dag}}\footnotemark[1]\textnormal{,} 
  \textbf{Yinghao Li}\textsuperscript{\textnormal{\dag}}\textnormal{,} \\
  \textbf{Dingu Sagar V K}\textsuperscript{\dag}, 
  \textbf{Rongzhi Zhang}\textsuperscript{\dag}, 
  \textbf{Changhao Li}\textsuperscript{\dag}, 
  \textbf{Ian Shu-Hei Wong}\textsuperscript{\dag}, \\
  \textbf{Sherry Yang}\textsuperscript{\S}, 
  \textbf{Percy Liang}\textsuperscript{\S}, 
  \textbf{Chao Zhang}\textsuperscript{\dag}, 
  \textbf{Bo Dai}\textsuperscript{\dag} \\
  \textsuperscript{\dag}Georgia Institute of Technology \quad
  \textsuperscript{\S}Stanford University
}

\begin{document}

% % %===================distance settings===============

% \setlength{\abovedisplayskip}{1pt}

% \setlength{\abovedisplayshortskip}{1pt}

% \setlength{\belowdisplayskip}{1pt}

% \setlength{\belowdisplayshortskip}{1pt}

% \setlength{\jot}{1pt}
 
% \setlength{\floatsep}{1ex}

% % \setlength{\textfloatsep}{1ex}

\newcolumntype{A}{>{\columncolor{blue!10}}c}   % AUP columns
\newcolumntype{H}{>{\columncolor{green!10}}c}  % H‑Rank (%) columns
\newcolumntype{E}{>{\columncolor{red!10}}c}    % Elo columns

\maketitle

\begin{abstract}
We introduce \method, a Gym-style framework for systematically reinforcement learning, evaluating, and improving autonomous large language model (LLM) agents in iterative machine learning engineering (MLE) workflows. 
Unlike existing benchmarks that primarily rely on static datasets or single-attempt evaluations, \method provides an interactive environment enabling agents to iteratively experiment, debug, and refine solutions through structured feedback loops. 
Built upon 200+ real-world Kaggle challenges. \method covers diverse, open-ended MLE tasks carefully curated to reflect realistic engineering scenarios such as data processing, architecture search, hyperparameter tuning, and code debugging.
Its fully executable environment supports comprehensive agent training via both supervised fine-tuning and reinforcement learning, facilitating iterative experimentation, realistic data sampling, and real-time outcome verification. 
Extensive evaluations of eight frontier LLMs reveal that while current models achieve meaningful iterative improvements, they still exhibit significant limitations in autonomously generating long-horizon solutions and efficiently resolving complex errors. 
Furthermore, \method's flexible and extensible architecture seamlessly integrates diverse data sources, tools, and evaluation protocols, uniquely enabling model-based agent tuning and promoting interoperability, scalability, and reproducibility. 
We open-source our framework and benchmarks to foster community-driven innovation towards next-generation MLE agents: \url{https://github.com/MLE-Dojo/MLE-Dojo}.
\footnote{\method webpage and leaderboard available at \url{https://mle-dojo.github.io/MLE-Dojo-page/} and \url{https://huggingface.co/spaces/MLE-Dojo/Leaderboard}}
\end{abstract}

\vspace{-1em}

% \begin{figure}[h]
%     \centering
%     \hspace*{-2em}
%     \includegraphics[width=0.95\linewidth]{figures/barplot/elo_categories_barplot3.pdf}
%     \caption{
%     % \Bo{wrap this figure.}
%     % Benchmark evaluations of eight frontiers LLMs across 50 evaluation tasks in \method.
%     Benchmark evaluations of eight frontiers LLMs across 50 evaluation tasks covering four main MLE domains in \method: Tabular, CV, NLP, and MLE-Lite.
%     % : MLE-Lite (a lightweight and diverse subset of MLE-Bench~\cite{chan2024mle}), tabular analysis (Tabular), natural language processing (NLP), and computer vision (CV).
%     % \Bo{the color of the bars here are too similar. please consider to use different colors.}
%     }
%     \label{fig:overall-bar}
% \end{figure}

% \begin{figure}
%     \centering
%     \hspace*{-2em}
%     \includegraphics[width=0.95\linewidth]{figures/barplot/elo_horizontal_barplot_last.pdf}
%     \caption{Caption}
%     \label{fig:enter-label}
% \end{figure}
\begin{wrapfigure}{r}{0.5\linewidth}
    \centering
    \hspace{-1em}
    \vspace{-2ex}
    \includegraphics[width=\linewidth]{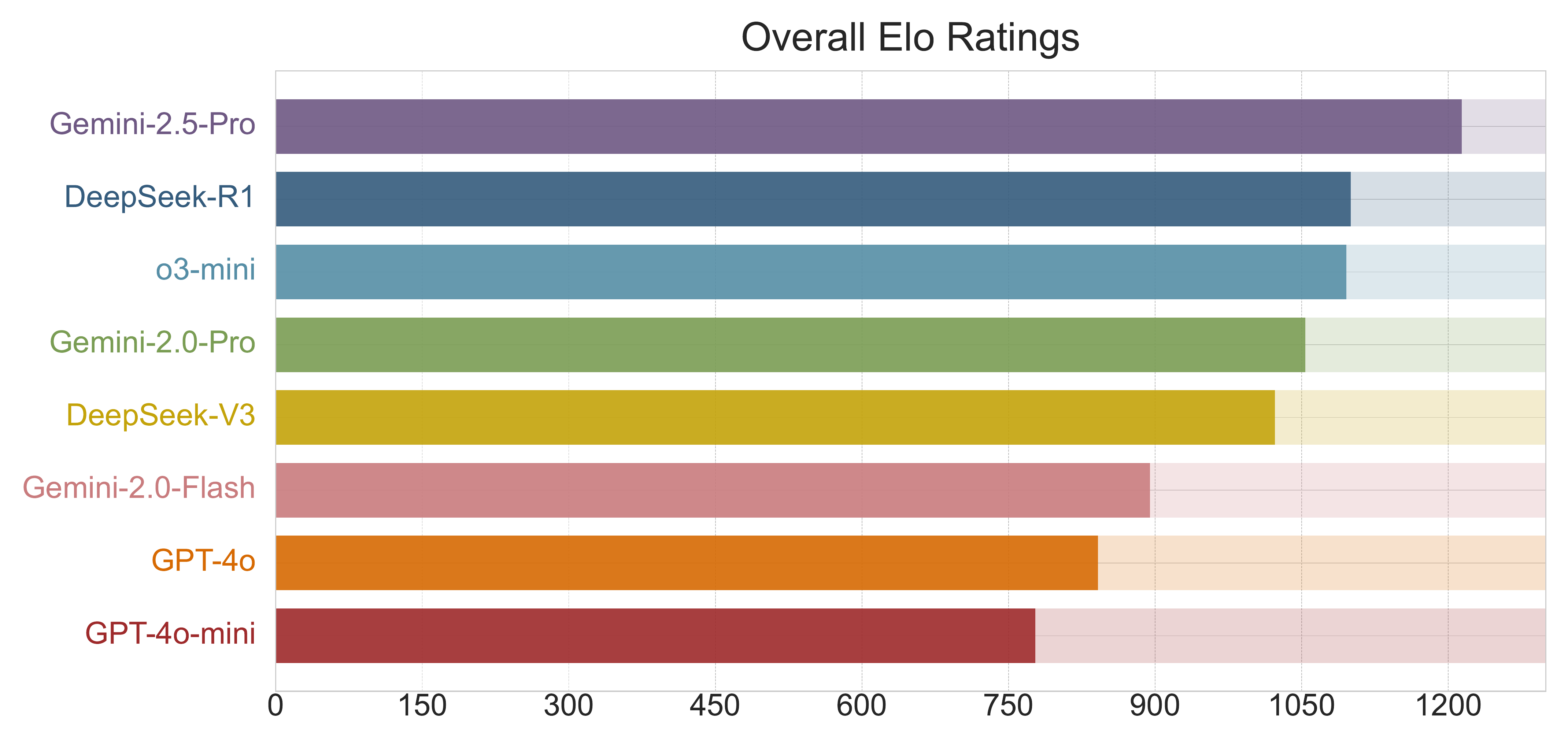}
    % \vspace{-1ex}
    \caption{Benchmark evaluations of eight frontiers LLMs across 50 evaluation tasks in \method.}
    \label{fig:history-radar}
    % \vspace{-1ex}
\end{wrapfigure}

% \begin{figure*}[h]
%     \centering
%     \begin{subfigure}
%         \centering
%         \includegraphics[width=0.4\textwidth]{figures/position_score/position_score_radar.pdf}
%         \label{fig:pos_score}
%     \end{subfigure}
%     \begin{subfigure}
%         \centering
%         \includegraphics[width=0.5\textwidth]{figures/elo/combined_elo_radar.pdf}
%         \label{fig:elo-radar}
%     \end{subfigure}
%     \caption{Benchmark evaluations of eight frontiers LLMs on our proposed \method across four main MLE domains in \method: tabular analysis, computer vision (CV), natural language processing (NLP), and MLE-Lite. \todo{figure to modify; not sure which term (MLE-Lite or Time series) you want to use in radar plot}. }
% \end{figure*}

% \begin{figure}[h]
%   \centering
%     % \vspace{-4ex}
%   \includegraphics[width=0.4\linewidth]{figures/position_score/position_score_radar.pdf}
%   \caption{
%   Position score radar figure. \todo{wrap figure}
%   }
%   % \vspace{-2ex}
%   \label{fig:pos_score}
% \end{figure}

% \begin{figure}[h]
%     \centering
%     \includegraphics[width=0.5\linewidth]{figures/elo/combined_elo_radar.pdf}
%     \caption{Performance of Elo rank}
%     \label{fig:elo-radar}
% \end{figure}

\section{Introduction}
\label{sec:intro}
Large language models (LLMs) have demonstrated remarkable capabilities across diverse coding tasks, including code generation, debugging, and refactoring~\cite{chen2021evaluating, hendrycks2021measuring, austin2021program, li2022competition}. Despite these advances, machine learning engineering (MLE) tasks remain uniquely challenging due to their inherent complexity, specialized domain knowledge, and the extensive iterative experimentation required for developing and optimizing ML algorithms~\cite{huang2024mlagentbench,nathani2025mlgymnewframeworkbenchmark,chan2024mle}. 
LLM-based agents hold significant promise for revolutionizing MLE by automating repetitive steps, generating boilerplate code, suggesting suitable algorithms, debugging implementations, and iteratively refining models. 
Ultimately, advanced LLM-based MLE agents could autonomously perform tasks such as literature search, hypothesis generation, experimental design, method implementation, result analysis, and even scientific dissemination. However, developing robust and autonomous MLE agents remains in its infancy, largely due to the absence of comprehensive benchmarks, interactive and executable training environments, and standardized evaluation frameworks capable of supporting rigorous, iterative experimentation. 
% \ycz{any better reasons: comprehensive tasks, accessible, executive, interactive environment? Data sampling trajectories? maybe can learn from mlgym: empirical validation, rigorous evaluation, and standardized benchmarks to ensure the reliability and reproducibility of findings; lack comprehensive frameworks and benchmarks specifically designed to assess their capabilities in conducting open-ended AI research tasks in diverse domains. This absence of standardized evaluation tools hinders our ability to objectively measure progress and identify areas for improvement in this emerging field. However, existing benchmarks for AI Research Agents either do not include open-ended research tasks, or only cover a narrow range of research domains. In addition, existing frameworks are not designed to enable research on different training algorithms for AI Research Agents such as reinforcement learning, curriculum learning, or open-ended learning. Finally, current frameworks do not allow flexible artifacts to be evaluated (e.g. different outputs of the agent's research such as a model, algorithm, or set of predictions).}

Several recent benchmarks have emerged to assess and facilitate progress in LLM-based coding agents~\cite{lai2023ds, yin2023natural, zhang2024pybench,huang2024mlagentbench,jing2025dsbench,nathani2025mlgymnewframeworkbenchmark,schmidgall2025agentrxiv}. However, most existing efforts focus mainly on isolated tasks such as data analysis or visualization, or competitions with narrow, non-interactive scenarios~\cite{lai2023ds, yin2023natural, zhang2024pybench}. Such compartmentalized benchmarks fail to capture the inherent complexity and iterative nature of real-world MLE workflows, which require continuous experimentation, iterative debugging, structured feedback incorporation, and efficient resource management. 
Although broader benchmarks such as MLE-Bench~\cite{huang2024mlagentbench} and DSBench~\cite{jing2025dsbench} provide more diverse MLE tasks, they still lack interactive environments that support iterative experimentation and training paradigms such as fine-tuning or reinforcement learning. 
Moreover, in contrast to structured software engineering (SWE) tasks~\cite{jimenez2023swe,zan2025multi}, which only require a structured environment for code-based problem-solving and can be readily sourced from general web resources such as GitHub issues, typical MLE tasks necessitate systematically curated datasets and standardized training data, elements currently unavailable in existing MLE benchmarks or gym-style frameworks~\cite{chan2024mle, nathani2025mlgymnewframeworkbenchmark}. 
This complexity not only increases storage and computational requirements, but also creates significant challenges when scaling to more comprehensive and diverse problem sets. 

In this study, we introduce \method, an interactive, large-scale Gym-style environment explicitly designed for developing and benchmarking autonomous LLM agents on real-world MLE workflows. Built upon 200+ real-world Kaggle competitions across critical ML domains, including tabular data analysis, computer vision, natural language processing, and time series forecasting, \etc. \method provides carefully curated, executable tasks that closely replicate real-world engineering challenges. Among these, 150 tasks constitute the initial training dataset for MLE agents and are integrated into an interactive environment, enabling training trajectory sampling for both supervised fine-tuning and reinforcement learning. Moreover, \method incorporates pre-installed dependencies, standardized evaluation scripts, and real-time outcome verification, thus simplifying iterative experimentation and debugging processes.
We conduct extensive empirical evaluations involving eight LLMs and multiple agent scaffolds. All evaluation results are publicly accessible via a continuously updated leaderboard, promoting transparent comparison, reproducibility, and collaborative research in this emerging field. Additionally, \method features a modular architecture that decouples agent capabilities from the underlying environment, facilitating seamless integration with diverse tools and data sources, enhancing interoperability, scalability, and supporting a robust ecosystem for developing generalizable MLE agents.
Our main contributions are summarized as follows:
\begin{itemize}[leftmargin=6mm]
    \item \textbf{Comprehensive Framework and Benchmark}: We establish \method as a comprehensive and large-scale benchmark consisting of over 200 Kaggle MLE competitions, enabling systematic and rigorous evaluations of autonomous LLM agents.

    \item \textbf{Interactive and Executable Environment}: \method provides an interactive and fully executable Gym-style environment that facilitates iterative experimentation, including comprehensive training trajectories sampling for supervised fine-tuning and reinforcement learning.

    \item \textbf{Advanced Functionalities and Scalability Supports}: \method uniquely facilitates outcome verification, model-agnostic agent tuning, and seamless integration of diverse datasets and tools, significantly accelerating the development of robust, generalizable, and scalable MLE agents.

    \item \textbf{Extensive Empirical Evaluation and Public Leaderboard}: We conduct large-scale evaluations across multiple state-of-the-art LLMs and agent scaffolds, with results publicly available through an actively maintained long-term real-time leaderboard to foster community-driven innovation.
\end{itemize}

In the remainder of the paper, we discuss the related work in \cref{sec:related works}, detail our data curation process in \cref{sec:data}, and thoroughly describe \method in \cref{sec:method}. We present the experimental setup and results of LLMs as MLE agents in \cref{sec:agent} and conclude the paper in \cref{sec:conclusion}.

\section{Related Works}
\label{sec:related works}

\begin{table*}[t]
\begin{threeparttable}
\centering 
\renewcommand\arraystretch{0.98}
\fontsize{7.5}{9.5}\selectfont \setlength{\tabcolsep}{0.3em}
% \vspace{-6ex}
\begin{tabular}{@{}l|ccccc|ccc|cc@{}}
\toprule
\textbf{Datasets} & \begin{tabular}[c]{@{}c@{}}\textbf{Interactive}\\ \textbf{Gym}\end{tabular} & \begin{tabular}[c]{@{}c@{}}\textbf{Executable}\\ \textbf{Environment}\end{tabular} & \begin{tabular}[c]{@{}c@{}}\textbf{Training}\\ \textbf{Facilities}\end{tabular} & \begin{tabular}[c]{@{}c@{}}\textbf{Extensible}\\ \textbf{Tasks}\end{tabular}& \begin{tabular}[c]{@{}c@{}}\textbf{Flexible}\\ \textbf{Scaffolds}\end{tabular}& \textbf{Tabular} & \textbf{CV}  & \textbf{NLP} & \textbf{\# Training} & \textbf{\# Eval}\\ \midrule 
AutoKaggle~\cite{li2024autokaggle} & \textcolor{red}{\XSolidBrush} & \textcolor{red}{\XSolidBrush} & \textcolor{red}{\XSolidBrush} & \textcolor{red}{\XSolidBrush} & \textcolor{red}{\XSolidBrush} & \textcolor{green}{\CheckmarkBold} & \textcolor{red}{\XSolidBrush} & \textcolor{red}{\XSolidBrush} & 0 & 8 \\
MLAgentBench~\cite{huang2024mlagentbench} & \textcolor{red}{\XSolidBrush} & \textcolor{red}{\XSolidBrush} & \textcolor{red}{\XSolidBrush} & \textcolor{red}{\XSolidBrush} & \textcolor{red}{\XSolidBrush} & \textcolor{green}{\CheckmarkBold} & \textcolor{green}{\CheckmarkBold} & \textcolor{green}{\CheckmarkBold} & 0 & 13 \\
DSBench~\cite{jing2025dsbench} & \textcolor{red}{\XSolidBrush} & \textcolor{red}{\XSolidBrush} & \textcolor{red}{\XSolidBrush} & \textcolor{red}{\XSolidBrush} & \textcolor{red}{\XSolidBrush} & \textcolor{green}{\CheckmarkBold} & \textcolor{green}{\CheckmarkBold} & \textcolor{green}{\CheckmarkBold} & 0 & 74 \\
DSEval~\cite{zhang2024benchmarking} & \textcolor{red}{\XSolidBrush} & \textcolor{green}{\CheckmarkBold} & \textcolor{red}{\XSolidBrush} & \textcolor{red}{\XSolidBrush} & \textcolor{red}{\XSolidBrush} & \textcolor{green}{\CheckmarkBold} & \textcolor{green}{\CheckmarkBold} & \textcolor{green}{\CheckmarkBold} & 0 & 31 \\
MLEBench~\cite{chan2024mle} & \textcolor{red}{\XSolidBrush} & \textcolor{green}{\CheckmarkBold} & \textcolor{red}{\XSolidBrush} & \textcolor{red}{\XSolidBrush} & \textcolor{green}{\CheckmarkBold} & \textcolor{green}{\CheckmarkBold} & \textcolor{green}{\CheckmarkBold} & \textcolor{green}{\CheckmarkBold} & 0 & 75 \\
MLGym~\cite{nathani2025mlgymnewframeworkbenchmark} & \textcolor{green}{\CheckmarkBold} & \textcolor{green}{\CheckmarkBold} & \textcolor{red}{\XSolidBrush} & \textcolor{red}{\XSolidBrush} & \textcolor{red}{\XSolidBrush} & \textcolor{green}{\CheckmarkBold} & \textcolor{green}{\CheckmarkBold} & \textcolor{green}{\CheckmarkBold} & 0 & 13 \\
\rowcolor{teal!12} \method & \textcolor{green}{\CheckmarkBold} & \textcolor{green}{\CheckmarkBold} & \textcolor{green}{\CheckmarkBold} & \textcolor{green}{\CheckmarkBold} & \textcolor{green}{\CheckmarkBold} & \textcolor{green}{\CheckmarkBold} & \textcolor{green}{\CheckmarkBold} & \textcolor{green}{\CheckmarkBold} & 150 & 50\\
\bottomrule
\end{tabular}
% \begin{tablenotes}
%     \item $^*$ In addition, the StarCoder-API can offer 4.77M more APIs.
% \end{tablenotes}
\caption{Summary of existing benchmarks of MLE agents with data resources and sample sizes. }
% \vspace{-2ex}
\label{tab:related}
\end{threeparttable}
\end{table*}

\textbf{\texttt{X}-Agent (Coding).}
LLM-based coding agents have significantly advanced automated software development, with impactful applications across two primary domains: SWE and MLE. 
\textbf{SWE agents}~\cite{yang2024swe,fourney2024magentic,wang2024openhands} excel at code generation, debugging, and refactoring for improved maintainability and performance. 
For example, SWE-agent~\cite{yang2024swe} operates within a constrained agent-computer interface to facilitate file creation, repository navigation, and code testing;
Magentic-One~\cite{fourney2024magentic} expands the capability of SWE agents with web navigation features, further enhancing their practicality and usability.
OpenHands~\cite{wang2024openhands} introduces an SWE agent with sandboxed environments to ensure safe command execution and verifiable web browsing, facilitating standardized benchmarking.
\textbf{MLE agents}~\cite{hollmann2024large,guo2024dsagent,li2024autokaggle,grosnit2024large,hong2024data,chi2024sela,jiang2025aide,trirat2024automl}, on the other hand, specialize in constructing, optimizing, and deploying comprehensive ML pipelines, including automatic architecture selection and hyperparameter tuning tailored to specific data characteristics. 
Early contributions focused on automating individual steps in ML engineering pipelines; for instance, CAAFE~\cite{hollmann2024large} leverages LLMs to iteratively generate semantically meaningful features for tabular data based on dataset description.
More recent works have progressively expanded automation from single steps to encompass end-to-end ML workflows.
For example, DS-Agent~\cite{guo2024dsagent} integrates a case-based reasoning approach to retrieve, reuse, evaluate, and refine solutions based on historical experiences. 
AutoKaggle~\cite{li2024autokaggle} develops iterative code execution, debugging, and comprehensive unit testing to ensure code correctness and logic consistency.
Agent K v1.0 \cite{grosnit2024large} provides an end-to-end autonomous data science agent designed to optimize workflows across diverse data science scenarios, while DataInterpreter \cite{hong2024data} employs hierarchical graph modeling for complex problem decomposition and programmable node generation that refines and verifies subproblems via code generation.
Advanced search and optimization approaches have further enhanced the capabilities of MLE agents.
SELA~\cite{chi2024sela} employs Monte Carlo Tree Search to enhance strategic planning during automated ML task-solving.
AIDE \cite{jiang2025aide} frames MLE as a code optimization problem, formulating trial-and-error as a tree search in the space of potential solutions.
Here, we focus on evaluating and improving MLE agents.

\textbf{\texttt{X}-Bench.}
Benchmarks for MLE agents not only provide foundational baselines but also inaugurate public leaderboards.
As illustrated in Table~\ref{tab:related}, early benchmarks primarily targeted isolated actions, such as data visualization and simple data manipulation~\cite{lai2023ds, yin2023natural, zhang2024pybench}. 
Recent benchmarks expand the scope of evaluations to encompass broader project-level tasks, often sourced from real-world Kaggle competitions. 
For example, AutoKaggle~\cite{li2024autokaggle} assesses static LLM workflows across 8 tabular Kaggle competitions. 
MLAgentBench \cite{huang2024mlagentbench} incorporates 13 tasks from Kaggle and bespoke ML challenges, providing baseline solutions and evaluating how often agents can achieve at least 10\% improvement over these baselines.
However, such benchmarks often exhibit limitations in both task breadth and complexity.
Benchmarks such as MLE-Bench~\cite{chan2024mle} and DSBench~\cite{jing2025dsbench} significantly broaden task diversity to 75 and 74 competitions, respectively, aiming to closely simulate real-world MLE scenarios. 
However, these platforms still lack robust interactive environments that support iterative experimentation, agent fine-tuning, and realistic training scenarios, critical components for effective development and evaluation of MLE agents. 
In contrast, \method combines comprehensive MLE task coverage with a fully interactive execution environment, offering unparalleled opportunities for developing, benchmarking, and refining advanced MLE agents.

\textbf{\texttt{X}-Gym (Dojo).} 
Interactive environments are crucial for evaluating and refining LLM-based agents. SWE-Gym~\cite{pan2024training} introduces an interactive environment specifically tailored for software engineering tasks, enabling extensive iterative training and validation. BrowserGym~\cite{drouin2024workarena} facilitates the evaluation of web-based navigation and interaction tasks, capturing complex real-world knowledge work scenarios. Similarly, Collaborative-Gym~\cite{shao2024collaborative} expands agent capabilities by enabling asynchronous, tripartite interaction among agents, humans, and task environments. 
The most recent and relevant work, ML-Gym~\cite{nathani2025mlgymnewframeworkbenchmark}, further integrates diverse ML research tasks into an interactive gym framework, providing an environment that supports reinforcement learning and iterative experimentation across open-ended research tasks. 
\method significantly extends both the quantity and complexity of tasks, offering 200+ competitions with an additional dedicated training set of 150 tasks. This enables effective trajectory sampling for supervised fine-tuning and reinforcement learning, thereby setting a new standard for comprehensive evaluation and training of MLE agents.

\section{Benchmark Tasks and Dataset Construction}
\label{sec:data}

\begin{figure*}[t]
  \centering
  \includegraphics[width=\linewidth]{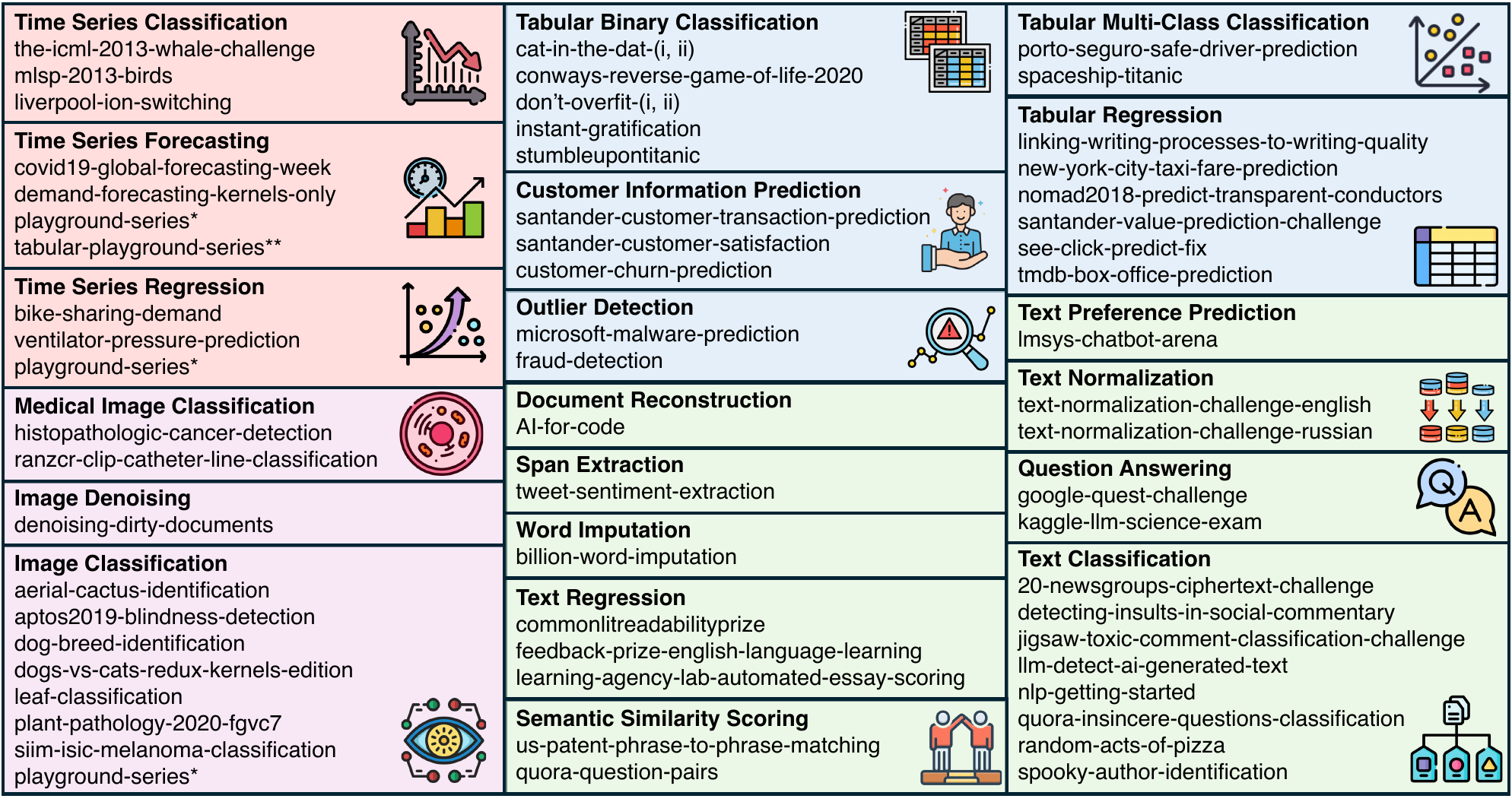}
  \caption{Overview of task diversity in \method, highlighting representative examples from four major domains: time series, computer vision, tabular data, and natural language processing. 
  }
  \label{fig:data}
\end{figure*}

\subsection{Task Description}
\method comprises more than 200 carefully selected tasks spanning diverse ML domains, including tabular data analysis, computer vision, natural language processing, and MLE-Lite\footnote{MLE-Lite is a lightweight yet comprehensive subset of MLE-Bench~\cite{chan2024mle}, encompassing 22 competitions across diverse domains, including CV, NLP, Tabular and Audio. It is constructed with a focus on both coverage and efficiency, and curated with user-friendliness inclusion in our evaluation suite.} (Figure~\ref{fig:data}). 
Each primary category further encompasses multiple specific task types---for instance, image classification within computer vision and sentiment analysis within natural language processing---yielding a total of 15 distinct task types.
The majority of these tasks represent complex, practical MLE challenges essential for developing robust LLM-based autonomous agents, sourced from real-world Kaggle competitions.
Kaggle\footnote{\url{https://www.kaggle.com/}} is a widely used online platform that hosts competitive ML tasks. Participants build ML models to solve defined challenges using real-world datasets, with performance publicly ranked based on predefined evaluation metrics. Leveraging extensive task library in Kaggle allows us to ensure task realism and broad relevance to practical MLE workflows.
% \yli{The rest is from...}

\subsection{Dataset Construction}
The objective of creating the dataset is to provide effective training data resources and solid evaluation over a diverse set of machine learning topics.
We select tasks that are light-weight tasks that are neither beginner nor too difficult and completed by a large number of human competitors.
This indicates that the selected tasks are representative and also relatively easy to validate and still testing the capabilities we are expecting from machine learning agents.
% \qrs{About train/test split: light-weight, medium difficult level(neither beginner nor too difficult, with relatively a large number of human competitors. We'll test more difficult level in the future.}

We aggregate tasks from multiple sources, including 68 competitions from MLE-Bench~\cite{chan2024mle} (excluding 7 tasks that are unavailable, excessively large, or tightly coupled with specific packages), 74 from DSBench~\cite{jing2025dsbench}, and 75 additionally scraped and prepared carefully from Kaggle's official website. After removing duplicate entries across sources, we obtain a diverse collection of over 200 unique tasks. 
To cover more aspects and data resources, we will continue to extend our dataset to include tasks with more complex data structures and higher difficulty, which require more sophisticated data processing and model training.
% \ycz{Not sure.}

\begin{wrapfigure}{r}{0.66\linewidth}
    \centering
    \vspace{-2ex}
    \includegraphics[width=\linewidth]{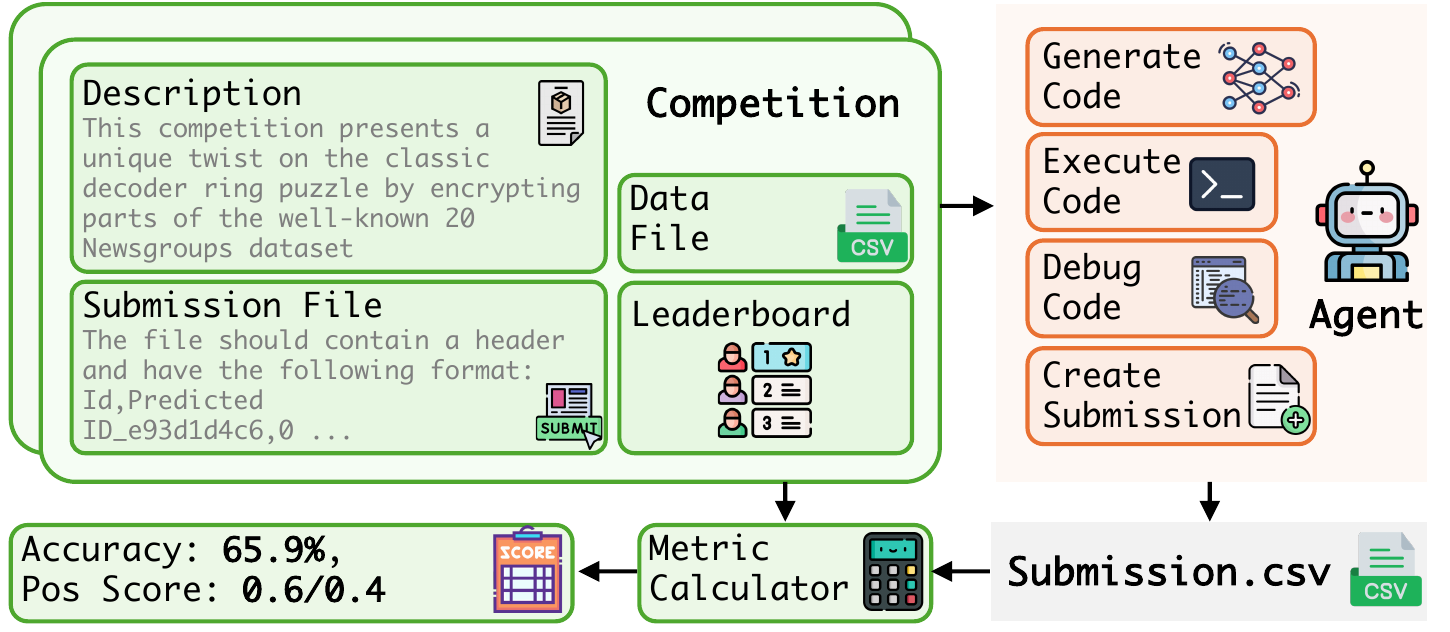}
  \caption{Overview of data structure in \method.
  }
    \label{fig:pipeline}
    \vspace{-1ex}
\end{wrapfigure}
Following MLE-Bench~\cite{chan2024mle}, each task is standardized into a consistent data structure (Figure~\ref{fig:pipeline}~and~\ref{fig:dir-structure}) to create a unified benchmark suitable for LLM agents, including:
(1) a \emph{detailed description} obtained from the competition website's ``Overview'' and ``Data'' sections;
(2) the \emph{original competition dataset}, reorganized into clear training and test splits; \footnote{Given Kaggle's potential absence of official test dataset answers.}
(3) an \emph{evaluation class} to locally measure performance against competition-specific metrics and validate submission format; and
(4) a \emph{snapshot of the competition leaderboard}, enabling performance comparisons (\ie, ranking) against human participants.
\method provides a standardized data interface and comprehensive documentation, allowing users to easily integrate new tasks or datasets into our structured, extensible format. 

We partition \method into training and evaluation subsets with a $150:50$ split. The evaluation set prioritizes tasks commonly referenced in prior benchmarks~\cite{chan2024mle,jing2025dsbench}, supplemented by strategically selected additional competitions to ensure domain diversity and representativeness. 
The larger training set provides extensive interactive experiences for LLM agents, facilitating robust agent training through both supervised fine-tuning and reinforcement learning.
% Tasks within \method fall under four primary categories: tabular data, computer vision, natural language processing, and time series analysis. \qrs{etc... } 

% \yli{This paragraph is a bit overlapping with 3.1.}

\section{\method}
\label{sec:method}

\subsection{Overview}

% \qrs{is it possible to include a section talking about the convenient apis and interface for users to use? showing how users can easily use our interface rather than others.}
% \yli{Shouldn't data construction and characteristics (section 3) be a part of this section?}

\begin{figure*}[t]
  \centering
    % \vspace{-4ex}
  \includegraphics[width=\linewidth]{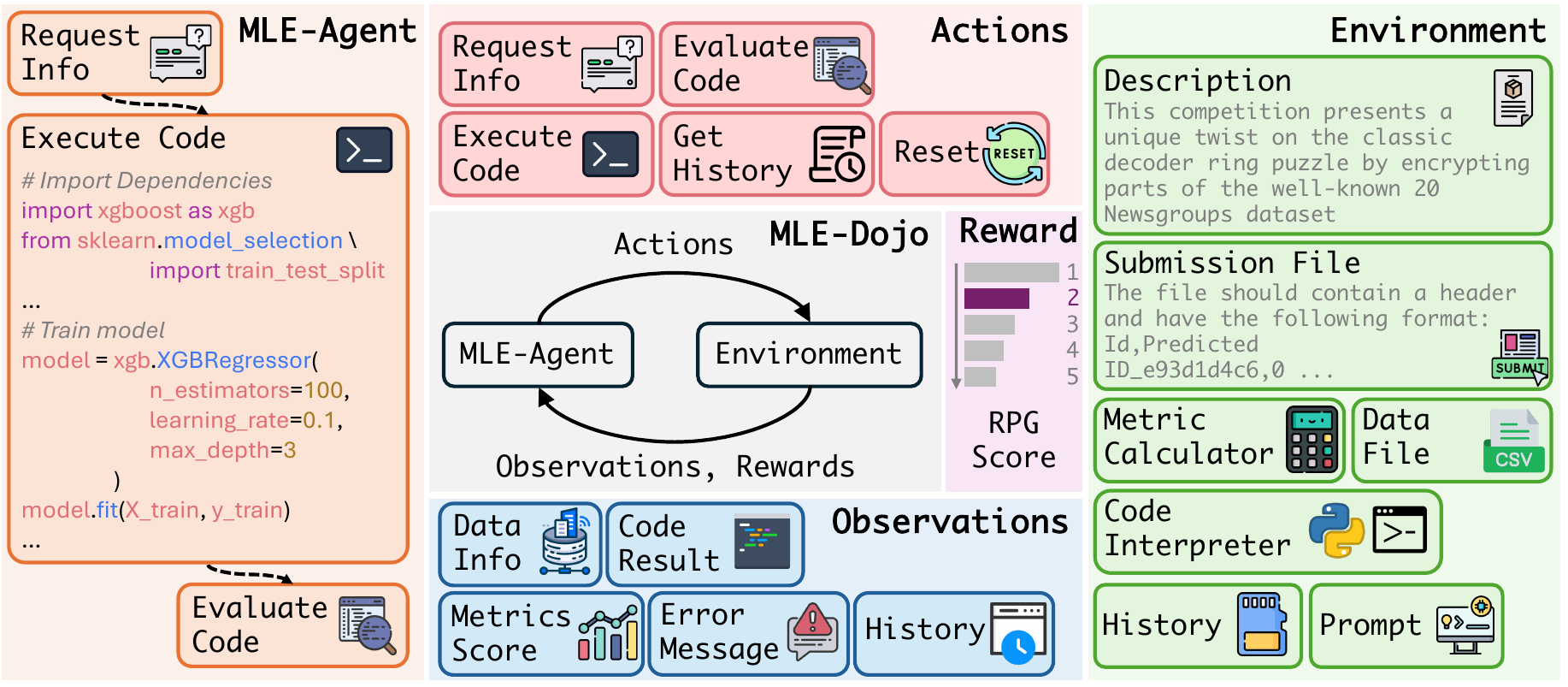}
  \caption{
  Overview of \method. The framework bridges MLE-Agents with MLE task environments through standardized interfaces for observation and action spaces.
  % \todo{\textbf{memory} --  we just offer a portal for the agent to connect}  \qrs{remove or change to another name like "history"}
  }
  % \vspace{-2ex}
  \label{fig:overview}
\end{figure*}

We propose \method, a benchmark that provides a standardized environment to evaluate MLE agents interacting with task-specific environments. 
As depicted in Figure~\ref{fig:overview}, \method provides a unified interface for MLE agents, \ie, LLM-based assistants that can write code to handle project-level ML tasks, such as completing data science competitions. 
Each task environment contains essential information, including datasets, evaluation metrics, analysis results, code execution outcomes, error messages, and interaction history, facilitating comprehensive agent-environment interactions.
The agent usually takes actions to solve the task, such as requesting task information, executing and writing code, evaluating generated code, retrieving past history, or resetting.

\begin{figure*}[t]
  \centering
    % \vspace{-4ex}
  \includegraphics[width=\linewidth]{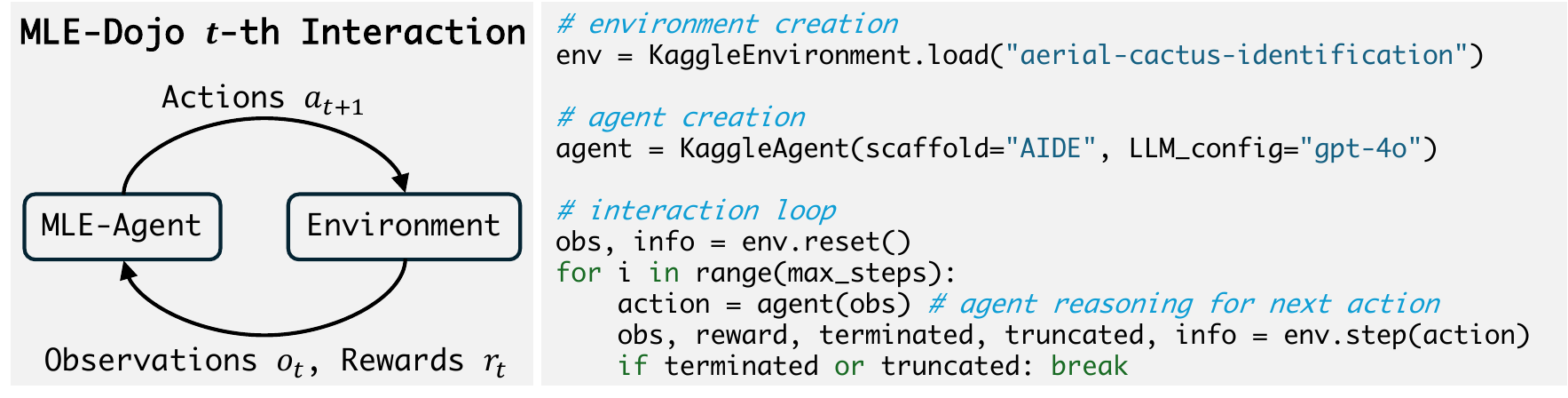}
  \caption{Interaction loop in \method with theoretical model (left) and concrete Python API (right). 
  % \footnote{For clarity and visual appeal, we present the core logic of the code in an intuitive manner. For detailed implementation, please refer to our implementation. }
  % \todo{@rushi: resolve this --> footnote not working }
}
  % \vspace{-2ex}
  \label{fig:code}
\end{figure*}

From an agent's perspective, the environment is typically centered on a sampled MLE problem $p\in\mathcal{P}$, where $\mathcal{P}$ is the task space. 
Each interaction between the environment and the agent can be formalized as a Partially Observable Markov Decision Process (POMDP), where at the time step $t$, the environment provides the current observation $o_t \in \mathcal{O}$ and the reward $r_t \in \mathcal{R}$. Depending on the specific agent and environment design, the agent generates the next action $a_{t+1} \in \mathcal{A}$ at step $t+1$ based on the history of prior interactions. Here, $\mathcal{O}$, $\mathcal{A}$, and $\mathcal{R}$ represent the observation space, action space, and reward space, respectively. The resulting interaction loop\footnote{For clarity and visual appeal, we present the core logic of the code in an intuitive manner. For detailed implementation, please refer to our code repository. } is shown in Figure~\ref{fig:code}.
% \qrs{take care of this part}
% \ycz{todo:add POMDP mathematical notations} 
% Formally, a POMDP is defined as a tuple $(\mathcal{S}, \mathcal{A}, \mathcal{T}, \mathcal{R}, \mathcal{O}, \mathcal{Z})$, where $\mathcal{S}$ is the state space, $\mathcal{A}$ is the action space, $\mathcal{T}: \mathcal{S} \times \mathcal{A} \times \mathcal{S} \rightarrow [0,1]$ is the transition function, $\mathcal{R}: \mathcal{S} \times \mathcal{A} \rightarrow \mathbb{R}$ is the reward function, $\mathcal{O}$ is the observation space, and $\mathcal{Z}: \mathcal{S} \times \mathcal{A} \times \mathcal{O} \rightarrow [0,1]$ is the observation function.

\subsection{Modular and User-Friendly Interface}
The \method framework is designed with a strong emphasis on modularity, flexibility, and extensibility. Each component within the environment operates both independently and collaboratively and is fully decoupled, allowing seamless integration and extension through unified \texttt{register} mechanism. Specifically, we define the following core modules, each of which encapsulates a distinct aspect of the environment:

\textbf{\texttt{Error}}. It encodes a comprehensive hierarchy of error types, enabling fine-grained debugging and facilitating informative feedback from the environment to the user.

\textbf{\texttt{Interface}}. It governs the execution and interaction logic of native environment actions, serving as the backbone of agent-environment communication.

\textbf{\texttt{Feedback}}. It translates interaction outcomes into structured, interpretable feedback to guide agent behavior and evaluation.

\textbf{\texttt{Metric}}. It defines a general metric base class, which can be subclassed to implement competition-specific evaluation metrics in a standardized and reusable manner.

Users can interact with the environment either by developing custom modules via the provided APIs or by directly using the pre-designed environment for seamless experimentation, requiring only a single call to \texttt{env.step(action\_type, action\_args)}. The minimal and intuitive interaction logic significantly lowers the barrier for agent customization and development. We have also integrated agents of several scaffolds into \method, as detailed in the Appendix~\ref{app:scaffolds}.
This design not only ensures clear separation of concerns, but also supports the easy incorporation of new functionalities, promoting both research reproducibility and rapid prototyping across diverse evaluation settings.

\subsection{Extensible Task Space $\mathcal{P}$}
The current task space encompasses all MLE tasks and competitions detailed in Section~\ref{sec:data}. Each task runs in a separate Docker container to isolate its execution environment. The dockerization approach ensures reproducibility and independence across tasks. Within each container, we implement a sandbox for executing agent-generated code. This sandbox is configurable with different experimental settings such as time limits and GPU/CPU memory constraints, providing a controllable, safe, and unified experimental environment. By separating the environment and the agent into private and public segments, \method ensures that certain directories remain inaccessible (private), while others are shared (public), offering a secure and reliable testing playground.

To simplify extending \method with new tasks, we standardize the data format for task integration (plug-and-play). 
A unified competition format is maintained for each task, including (1) \emph{Basic information}: a competition description and sample submission, (2) \emph{Well-structured datasets}: well-split datasets through a general \texttt{prepare.py} script (some competitions do not originally provide labeled test data), and (3) \emph{Public and private leaderboards}: both public and private leaderboards assessed using competition-specific evaluation metrics. Following this standardized data format, users can easily incorporate their own competitions for testing purposes. Comprehensive format definitions and detailed guidance are available in Appendix~\ref{app:structure}.

\subsection{Observation Space $\mathcal{O}$}
\method provides a rich observation space with five main components: 

\textbf{Dataset Information.} 
For each competition, \method supplies comprehensive information required to solve tasks, including (1) \emph{Competition background}: a concise overview of the historical context and evolving research challenges, 
(2) \emph{Goal description}: clear definition of specific objectives, desired outcomes, and evaluation metrics, 
(3) \emph{Sample submission}: templates exemplifying the expected format and content structure, and
(4) \emph{Data folder structure}: organization and naming conventions of provided datasets for user access and navigation. 

\textbf{Evaluation Metric Scores.}  
For each task, \method implements well-defined evaluation functions in Python, providing description-aligned criteria and metrics to validate submission format and method performance. Performance is reported as both \emph{raw scores} and \emph{HumanRank score}, offering absolute and relative performance insights. Additionally, \method provides detailed feedback on submission format to guide agents in actively modifying results accordingly.

\textbf{Code Execution Results.} 
To solve MLE tasks, agents generate Python code as solutions. For submission, code is executed in the sandbox environment, generating a submission file in a predefined directory. Once a submission file is successfully generated and follows the requirements specified in the Dataset Information, the environment calls the evaluation metric function to calculate results based on ground truth data and the submission file.

\textbf{Error Messages.} 
When code is executed in the sandbox, flaws may lead to \texttt{CompileError} or \texttt{RuntimeError}. To enable LLM agents to iteratively improve their generated code, \method encapsulates error messages and provides them during agent interactions. Even when code executes successfully, \method also reports additional error messages regarding submission files (\eg, incorrect formats, mismatching columns, \etc).

\textbf{Interaction History.} 
In \method, MLE agents can access interaction histories in two ways:
(1) \emph{Conversation history}: Obtained from the agent side, including all LLM generations and environment observations, and
(2) \emph{Environment records}: Obtained from the environment side, objectively recording every agent action and environment observation.
This dual approach accommodates different LLM agents and MLE task implementations, improving compatibility with both dialog-based LLMs and coding-focused agents. For reasoning models, conversation history effectively records extended reasoning procedures, while environment records better suit coding LLMs that may not support natural language interaction.

\subsection{Expandable Action Space $\mathcal{A}$}

The fundamental action space of \method consists of raw executable Python code.
Agents can call \texttt{request\_info} to query task descriptions and dataset details, then generate Python code to solve specific MLE tasks using three core functions: \texttt{validate\_code}, \texttt{execute\_code}, and \texttt{reset}, which connect the agent to the code interpreter in the sandbox. Additionally, agents can leverage the \texttt{get\_history} function to retrieve past interactions between the environment and themselves.
\method also provides a user-defined action portal, offering an \emph{extensible action space} for researchers to register additional actions. As long as new actions are registered through the portal and provided to the agent in context, the agent can learn to leverage these new actions to solve problems. By default, \method provides a predefined yet extensible set of five basic actions:

\textbf{\texttt{request\_info}.}
When an LLM agent requires necessary information to solve an MLE problem, it can call the \texttt{request\_info} function to access task descriptions, sample submissions, data directories, output directories, and data structures. The response includes all necessary competition information without further analysis or information pruning.

\textbf{\texttt{validate\_code}.}
The \texttt{validate\_code} action performs basic syntax checking and runtime validation. It provides a compilation trial of the code and returns detailed syntax or execution error messages. It also serves as a critical interface for running data analysis code and extracting deeper insights from the resulting outputs.  This action is more lightweight than \texttt{execute\_code} without submission requirements, which may serve as a debugging or information printing tool.

\textbf{\texttt{execute\_code}.}
Unlike \texttt{validate\_code}, the \texttt{execute\_code} action performs complete code execution, submission verification and evaluation. Only through this action can the generated submission of the code be validated and evaluated using the corresponding metric.
Each use of the \texttt{execute\_code} action corresponds to a full competition submission. Restriction on invocations of this action allow for different and flexible methodological and experimental design needs. Only through the invocation of this action can the generated submission be validated and evaluated using the corresponding metric.

\textbf{\texttt{get\_history}.}
Since memory is a crucial component in agent design, \method provides the \texttt{get\_history} function to access past experiences and enable learning from memory. This flexible design accommodates different agent architectures and scaffolding approaches.

\textbf{\texttt{reset}.}
\method offers a \texttt{reset} function to reset the entire environment and restart from scratch. This action serves as the default mechanism designed to accommodate a wide range of use cases.

\begin{figure}[t]
    \centering
    \includegraphics[width=0.9\linewidth]{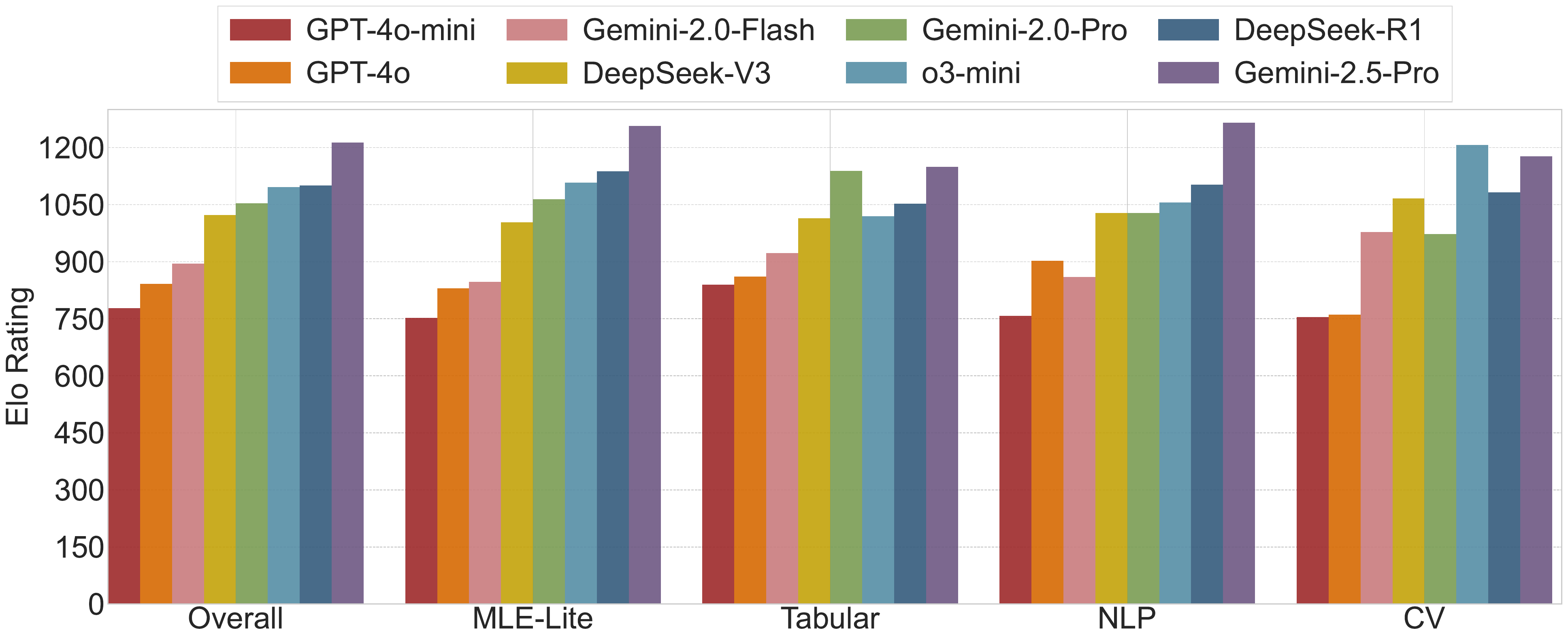}
    \caption{Elo Rankings of eight frontiers LLMs on our proposed \method across four main MLE domains in \method: tabular analysis, computer vision (CV), natural language processing (NLP), and MLE-Lite.}
    % \vspace{-1em}
    \label{fig:all-cat-bar}
\end{figure}

\subsection{Reward Space and Environmental Feedback}
\method introduces a reward mechanism specifically designed to reflect the quality of solutions for coding tasks. This mechanism utilizes quantitative metrics provided by each competition, such as accuracy, F1-score, etc. However, since competitions differ in evaluation metrics and score ranges, directly using absolute performance to evaluate LLM-based agents' generated code is not precise.
To provide continuous and fine-grained feedback that rewards incremental improvements and innovative solutions, we propose using the relative position of the human leaderboard as reward instead of the coarse-grained medals used in existing benchmarks~\cite{chan2024mle}.

\textbf{HumanRank Score.}
We calculate the relative position score of the current submission on the leaderboard of human competitors. Suppose that the submission ranks at position $p$ among a total of $N$ submissions on the leaderboard. Then, the position score is computed as: $s=1-\frac{p}{N}$\label{eq:ps}.
The HumanRank score indicates the percentage of human competitors the agent surpasses on the leaderboard for a given competition. A higher score reflects stronger performance relative to human participants.
To prevent bias between public and private leaderboards, we compute the relative scores on each leaderboard independently and then use their average as the final reward.

We adopt the HumanRank Score as the reward in our environment. First, it is fully aligned with the original performance metrics-achieving a higher original score is strictly positively correlated with obtaining a higher HumanRank Score, regardless of the specific metric used. Moreover, HumanRank is a normalized score within the [0, 1] range, which resolves the issue of varying score magnitudes across different tasks and enables it to serve as a unified and informative reward.
% \todo{align with evaluation}
% \qrs{HumanRank is our reward. Getting better raw scores will lead to higher HumanRank score, and since HumanRank is between 0-1, it provides unified value threshold for a reasonable, unified and meaningful reward.}
% \qrs{change a name with "human" involved, represent how many human competitors on the leaderboards are surpassed.}

% \textbf{ELO Ranking.}

\section{LLMs as MLE Agents in \method}
\label{sec:agent}

\subsection{Experiment Setups}

\textbf{Backbone LLMs.} 
We consider different backbones to test the effectiveness of \method in the evaluation and improvement of LLMs as MLE Agents. MLE Agent leverages native actions and interacts with the environment through a straightforward logic. Prompts and implementation details are available in appendix~\ref{app:prompt} and appendix~\ref{app:MLE-Agent}.
% \todo{need to introduce mle agent, the first time it occurs as our benchmarking scaffolds.} 
Specifically, we consider \texttt{gpt-4o-mini (2024-07-18)}~\cite{hurst2024gpt}, \texttt{gpt-4o (2024-11-20)}~\cite{hurst2024gpt}, \texttt{o3-mini (2025-01-31)}~\cite{o3-mini} from OpenAI, \texttt{Gemini-2.0-Flash}~\cite{gemini-2-flash}, \texttt{Gemini-2.0-Pro (exp)}~\cite{gemini-2-pro}, and \texttt{Gemini-2.5-Pro (exp-3-25)}~\cite{gemini-2.5-pro} from Google, and \texttt{DeepSeek-v3 (2025-03-24)}~\cite{liu2024deepseek} and \texttt{DeepSeek-r1}~\cite{deepseekai2025deepseekr1incentivizingreasoningcapability} from DeepSeek as evaluation backbone LLMs.
For non-reasoning models, we set temperature=$0.0$ and top-$p=1.0$ to ensure reproducible evaluations. We take the best performance of two runs per task per model.

\textbf{Evaluation Metrics.} 
To ensure a comprehensive evaluation, we consider \emph{Area Under the Performance Profile (AUP)}~\citep{nathani2025mlgymnewframeworkbenchmark}, \emph{HumanRank Score (H-Rank, \%)}, and \emph{Elo} ranking~\citep{chiang2024chatbot} together as metrics. Additional implementation details are available in \cref{app:implementation}.

\textbf{Environment Configurations.}
The total number of steps is set to 15, with agents having full access to their interaction histories. A unified, concise prompt with clear instructions is provided, without additional extraneous information. The maximum runtime per session is 12 hours, and GPU memory is limited to 32 GB. The maximum input token length is set at 50,000, while each output round is capped at 8,192 tokens. These prompt, history, time, and memory configurations are designed to rigorously evaluate LLMs' capabilities in long-context handling, instruction-following, reasoning, and coding under resource-constrained conditions, closely mirroring realistic Kaggle competition scenarios. We do not restrict the number of submission attempts to enable continuous improvement. To generate valid submission files and scores, agents must explicitly use the \texttt{"execute\_code"} command; submissions are not automated. We pre-install commonly used Python packages, though agents may install additional packages within their generated code as needed.

\subsection{Main Results}

\begin{table*}[t]
\centering
\fontsize{7}{9}\selectfont
\setlength{\tabcolsep}{0.2em}
\begin{tabular}{@{}l A H E  A H E  A H E  A H E @{}}
\toprule
\textbf{Task Categories ($\rightarrow$)} 
  & \multicolumn{3}{c}{\textbf{MLE-Lite}} 
  & \multicolumn{3}{c}{\textbf{Tabular}}
  & \multicolumn{3}{c}{\textbf{NLP}} 
  & \multicolumn{3}{c}{\textbf{CV}} \\
\cmidrule(lr){2-4} \cmidrule(lr){5-7} \cmidrule(lr){8-10} \cmidrule(lr){11-13}
\textbf{Models ($\downarrow$)} 
  & \textbf{AUP\textsuperscript{$\uparrow$}} & \textbf{H-Rank (\%)\textsuperscript{$\uparrow$}} & \textbf{Elo\textsuperscript{$\uparrow$}} 
  & \textbf{AUP\textsuperscript{$\uparrow$}} & \textbf{H-Rank (\%)\textsuperscript{$\uparrow$}} & \textbf{Elo\textsuperscript{$\uparrow$}} 
  & \textbf{AUP\textsuperscript{$\uparrow$}} & \textbf{H-Rank (\%)\textsuperscript{$\uparrow$}} & \textbf{Elo\textsuperscript{$\uparrow$}}  
  & \textbf{AUP\textsuperscript{$\uparrow$}} & \textbf{H-Rank (\%)\textsuperscript{$\uparrow$}} & \textbf{Elo\textsuperscript{$\uparrow$}}  \\
\midrule
gpt-4o-mini~\cite{hurst2024gpt} 
  & 1.492 & 21.21 & 753 
  & 0.724 & 15.37 & 839 
  & 0.837 & 13.14 & 758 
  & 1.172 & 10.17 & 754 \\
gpt-4o~\cite{hurst2024gpt} 
  & 1.448 & 27.85 & 830
  & 0.691 & 18.97 & 861
  & 0.842 & 29.97 & 903
  & 1.418 & 18.76 & 761 \\
o3-mini~\cite{o3-mini} 
  & 1.895 & 56.48 & 1108
  & 0.739 & 32.65 & 1019
  & 0.992 & 37.46 & 1056
  & 1.892 & 35.02 & 1207 \\
\midrule
DeepSeek-v3~\cite{liu2024deepseek} 
  & 1.825 & 44.26 & 1004 
  & 0.727 & 37.85 & 1015 
  & 0.977 & 28.41 & 1028 
  & 1.784 & 26.75 & 1067 \\
DeepSeek-r1~\cite{deepseekai2025deepseekr1incentivizingreasoningcapability} 
  & 1.852 & 58.43 & 1137
  & 0.678 & 38.13 & 1053
  & 0.988 & 28.48 & 1103
  & 1.844 & 34.26 & 1083 \\
\midrule
Gemini-2.0-Flash~\cite{gemini-2-flash} 
  & 1.696 & 33.50 & 847 
  & 0.689 & 30.36 & 923 
  & 0.884 & 28.39 & 860 
  & 1.607 & 20.35 & 978 \\
Gemini-2.0-Pro~\cite{gemini-2-pro} 
  & 1.796 & 48.61 & 1064
  & 0.787 & 37.46 & 1139
  & 0.970 & 30.93 & 1028
  & 1.651 & 23.07 & 973 \\
Gemini-2.5-Pro~\cite{gemini-2.5-pro} 
  & 1.919 & 61.95 & 1257 
  & 0.798 & 42.64 & 1150 
  & 0.998 & 38.45 & 1266 
  & 1.915 & 42.83 & 1177 \\
\bottomrule
\end{tabular}
\caption{Main experiments of LLMs as MLE Agents on MLE tasks in \method.}
\label{tab:mainexp}
\end{table*}

\begin{figure}[h]
    \centering
    \includegraphics[width=1.0\linewidth]{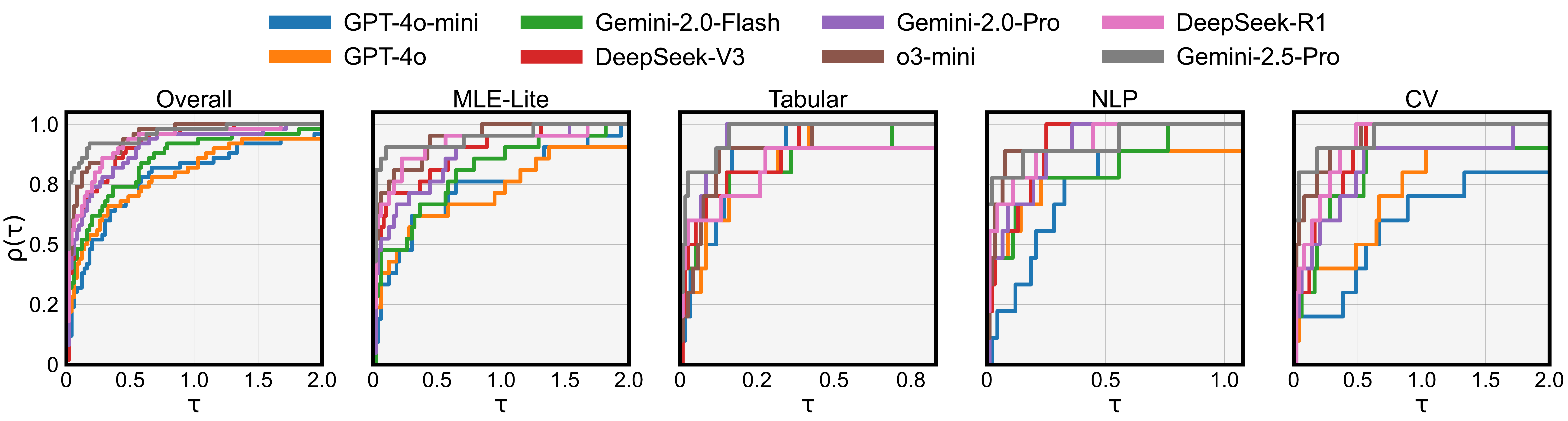}
    \caption{Performance profiles and corresponding AUP curves for evaluating the robustness of LLMs across four ML tasks. The x-axis represents the performance ratio threshold $\tau$, while the y-axis indicates the fraction of tasks for which a model achieves performance within a factor $\tau$ of the best-performing model.}
    \label{fig:aup}
\end{figure}

Table~\ref{tab:mainexp} presents a comprehensive evaluation of eight LLMs as MLE Agents across four fundamental ML tasks. 
% We have the following observations:
% \paragraph{Comparative Analysis of Model Performance.}
Reasoning and coding models such as \texttt{o3-mini}, \texttt{DeepSeek-r1}, and \texttt{Gemini-2.5-Pro} consistently achieve high rankings across all metrics, demonstrating strong adaptability, robustness, and overall effectiveness as MLE Agents. Additionally, Figure~\ref{fig:aup} further illustrates the Performance Profiles along with the corresponding AUP curves. 
Models like \texttt{Gemini-2.0-Pro} exhibit balanced performance profiles, achieving moderate but consistently reliable results across various tasks. This comprehensive comparative perspective highlights the strengths and limitations of each model, offering practical insights into their suitability for different MLE scenarios. 
In addition, the HumanRank Score provides an absolute measure of performance incorporating human benchmarks, the Elo Score clarifies competitive relationships through pairwise analyses, and the Performance Profiles with AUP scores assess robustness and consistency across performance variations. Together, these evaluation approaches in \method form a robust, multifaceted perspective on the capabilities and limitations of LLMs as MLE agents, enhancing the overall interpretation and reliability of our experimental conclusions. 

\begin{figure}[h]
    \centering
    % \vspace{-1ex}
    \includegraphics[width=1.0\linewidth]{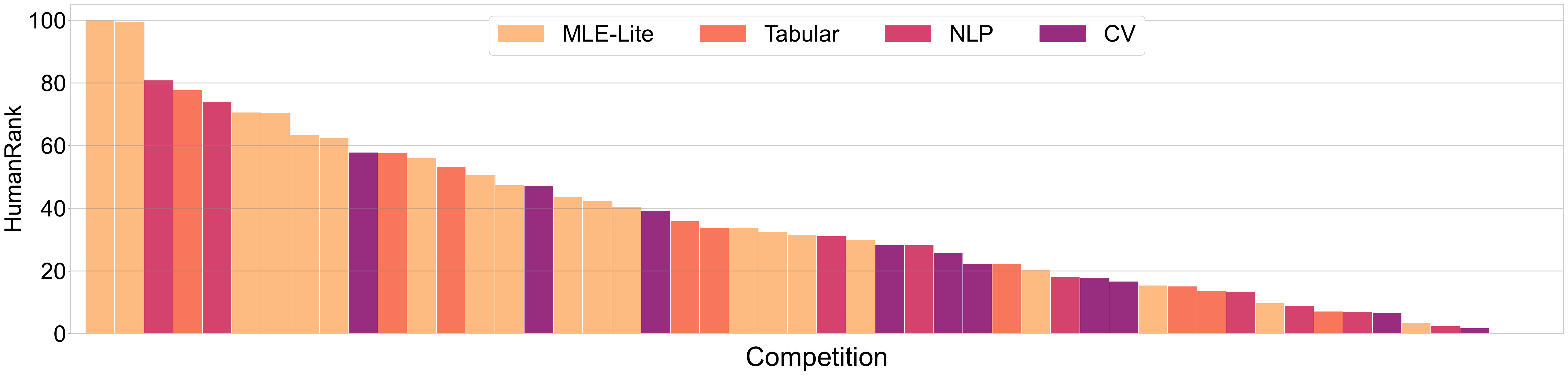}
    % \vspace{-1em}
    \caption{Task difficulty sorted by average HumanRank score.}
    % \vspace{-0.5em}
    \label{fig:difficulty}

\end{figure}

% \todo{@Rushi: draw an average performance task performance figure like in AIDE and place it here.}
Furthermore, we define the difficulty level of different tasks with the average performance of different models in comparison with the human leaderboard. 
Figure~\ref{fig:difficulty} illustrates the average performance distribution across 8 frontier models on the tasks.
As shown in the figure, CV tasks are the most challenging-none of them have an average HumanRank score above 60, and more than half fall below 30. For MLE-Lite tasks, the average HumanRank scores mostly exceed 30. Difficulty levels of tasks in other domains are more evenly distributed. 
% \todo{@Rushi: think about 1-2 observations/insights of the task difficulties, like ``from the figure, we get to know that the CV tasks are overall more difficult than tabular tasks?''}
% \qrs{list a table of difficulty based one HumanRank score? based on average across models}

\subsection{Cost Analysis}
\begin{figure}[t]
    \centering
    \includegraphics[width=1.0\linewidth]{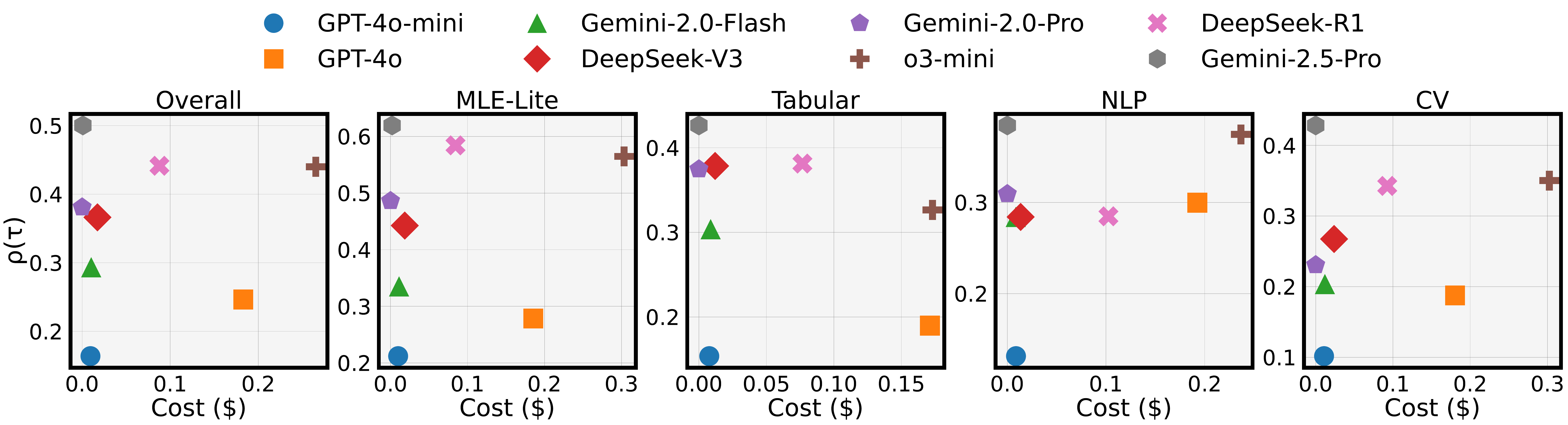}
    \caption{Relationship between average computational cost and performance across evaluated LLMs and task categories. Each point represents the average cost per task, with specific attention given to reasoning vs. non-reasoning model cost dynamics. Note that \texttt{Gemini-2.0-Pro} and \texttt{Gemini-2.5-Pro} are excluded from cost analysis (only for performance reference) due to current free usage.}
    
    \label{fig:perf-cost}
\end{figure}

Figure \ref{fig:perf-cost} illustrates the cost-performance relationship across different LLMs and task categories. 
Reasoning models (\eg, \texttt{DeepSeek-r1}) typically incur higher costs due to their premium pricing structures and longer solution outputs.
Even reasoning models with comparatively lower pricing, such as \texttt{o3-mini}, tend to produce longer outputs due to more complex reasoning processes. These longer outputs significantly increase overall token consumption, contributing to higher cumulative costs.
Notably, tasks involving computer vision and deep neural network training pipelines consistently generate longer codes compared to classical ML tasks (\eg, tabular analysis) executed on CPUs.
While cost generally correlates with solution complexity and token consumption, some models, such as \texttt{DeepSeek-r1}, achieve competitive performance with significantly fewer tokens, highlighting potential cost-efficiency opportunities.

\begin{figure}[t]
    \centering
    \includegraphics[width=1.0\linewidth]{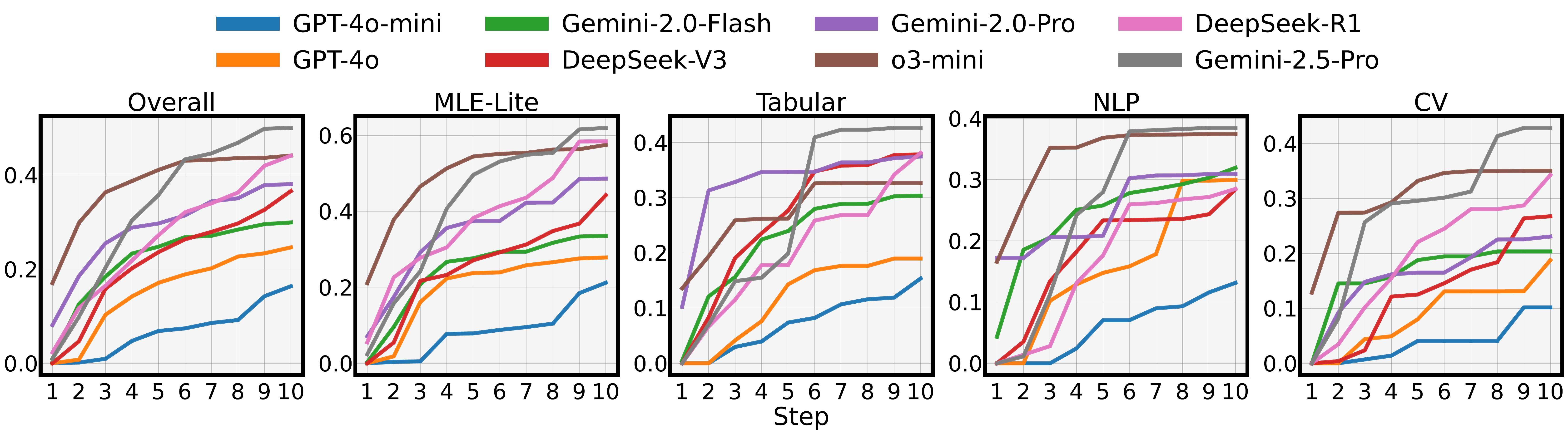}
    \caption{Step-wise HumanRank performance comparisons between reasoning and non-reasoning models, considering only code execution and validation steps. Information-requesting steps are excluded.}
    % \todo{which metrics using here?}
    \label{fig:perf-step}
\end{figure}

\subsection{Step-wise Performance Dynamics}
Figure \ref{fig:perf-step} presents the step-wise performance improvements across different models, illustrating variations in performance trajectories between reasoning and non-reasoning models.
Among reasoning models, \texttt{o3-mini} consistently achieves high performance within the initial steps (typically within the first five) and maintains stable scores in subsequent steps. Conversely, \texttt{DeepSeek-r1} and \texttt{Gemini-2.5-Pro} exhibit gradual improvements, achieving comparable or superior performance in intermediate to later steps.
Non-reasoning models occasionally outperform reasoning models at early or intermediate steps but generally show limited improvement as steps progress, resulting in lower final scores.

\subsection{Action and Error Analysis}
\begin{figure}[t]
\centering
\includegraphics[width=1.0\linewidth]{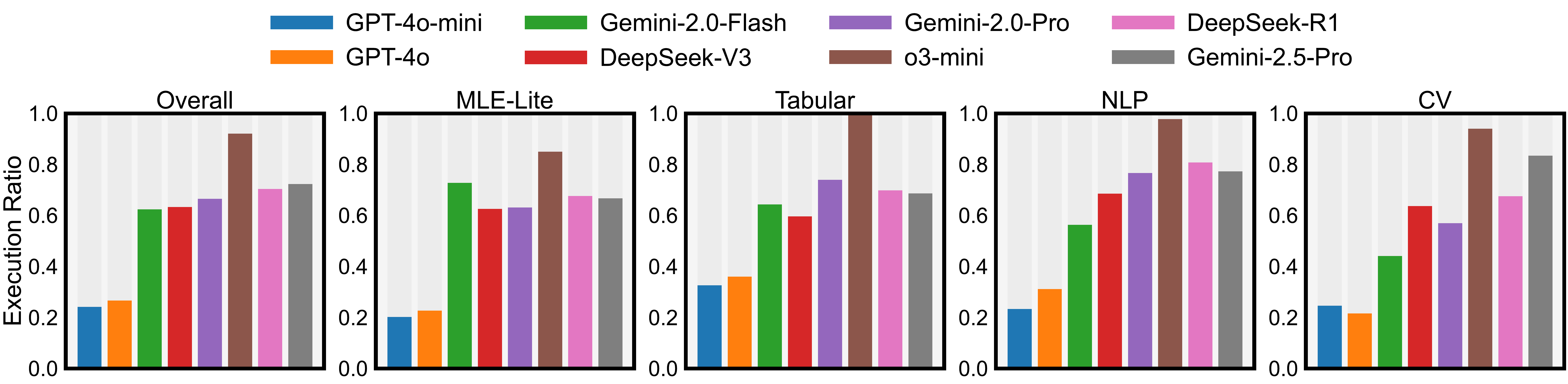}
\caption{Proportion of code execution actions relative to combined execution and validation actions.}
\label{fig:exe-ratio}
\end{figure}

\paragraph{Action Strategy and Execution Proportions.}
Models exhibit distinct strategies regarding the balance between code execution and validation actions (Figure \ref{fig:exe-ratio}). Specifically, \texttt{o3-mini} demonstrates high confidence by frequently proceeding directly to code execution without extensive preliminary validation or output inspections, adopting execution actions in over 90\% of instances. Conversely, \texttt{gpt-4o} and \texttt{gpt-4o-mini} employ significantly more conservative strategies, executing code only about 20\% of the time and relying heavily on validation steps. Notably, \texttt{Gemini-2.0-Flash}, despite not being among the top-performing models, uses an aggressive execution strategy comparable to stronger models.

\begin{wrapfigure}{r}{0.6\linewidth}
    \centering
    \vspace{-2ex}
    \includegraphics[width=\linewidth]{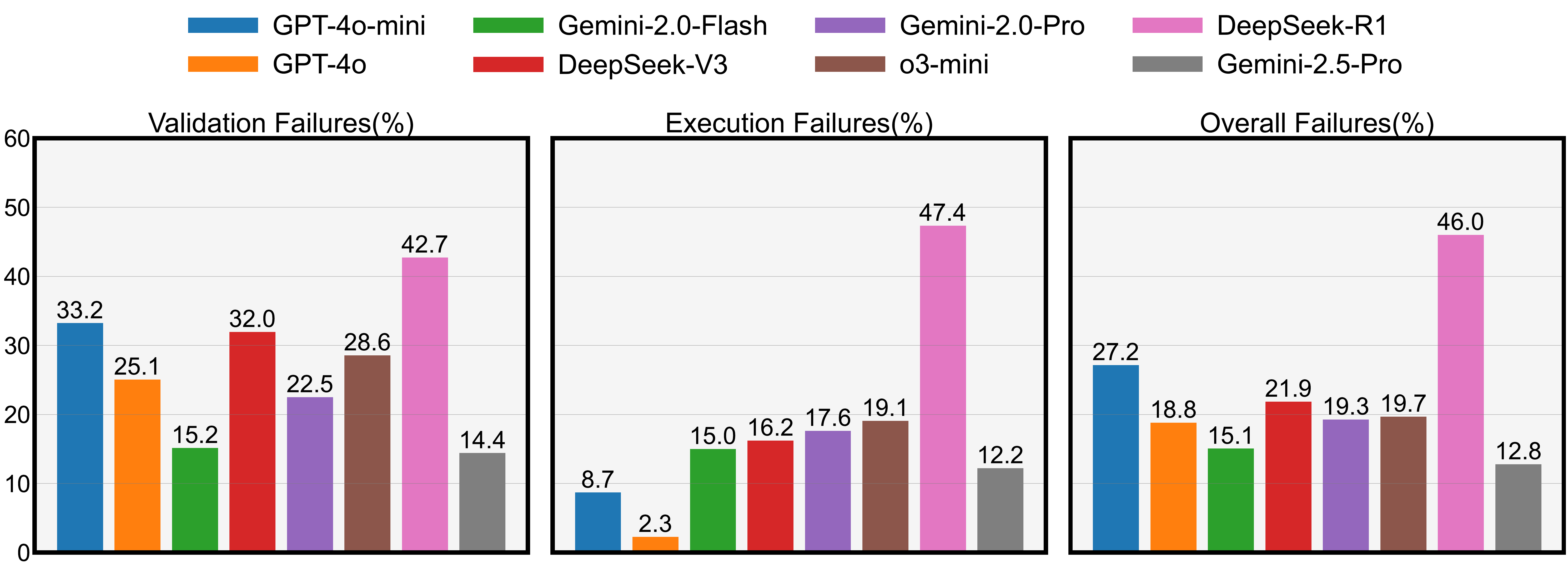}
  \caption{Average failure rates across tasks.
  }
    \label{fig:failure-rates}
    \vspace{-1ex}
\end{wrapfigure}
\paragraph{Failure Rate Analysis.}
Figure \ref{fig:failure-rates} shows the average failure rates across tasks, highlighting validation, execution, and overall errors. \texttt{Gemini-2.5-Pro} maintains the lowest overall failure rate, aligning with its consistently high performance. In contrast, \texttt{DeepSeek-r1}, despite achieving strong performance, experiences relatively high failure rates in both execution and validation categories. \texttt{gpt-4o} and \texttt{gpt-4o-mini}, due to their conservative strategies, achieve low execution failure rates but face comparatively high validation failures. Conversely, \texttt{Gemini-2.0-Flash} successfully balances aggressive execution with one of the lowest overall failure rates, second only to \texttt{Gemini-2.5-Pro}.

\begin{wrapfigure}{r}{0.4\linewidth}
    \centering
    \vspace{-1ex}
    \includegraphics[width=\linewidth]{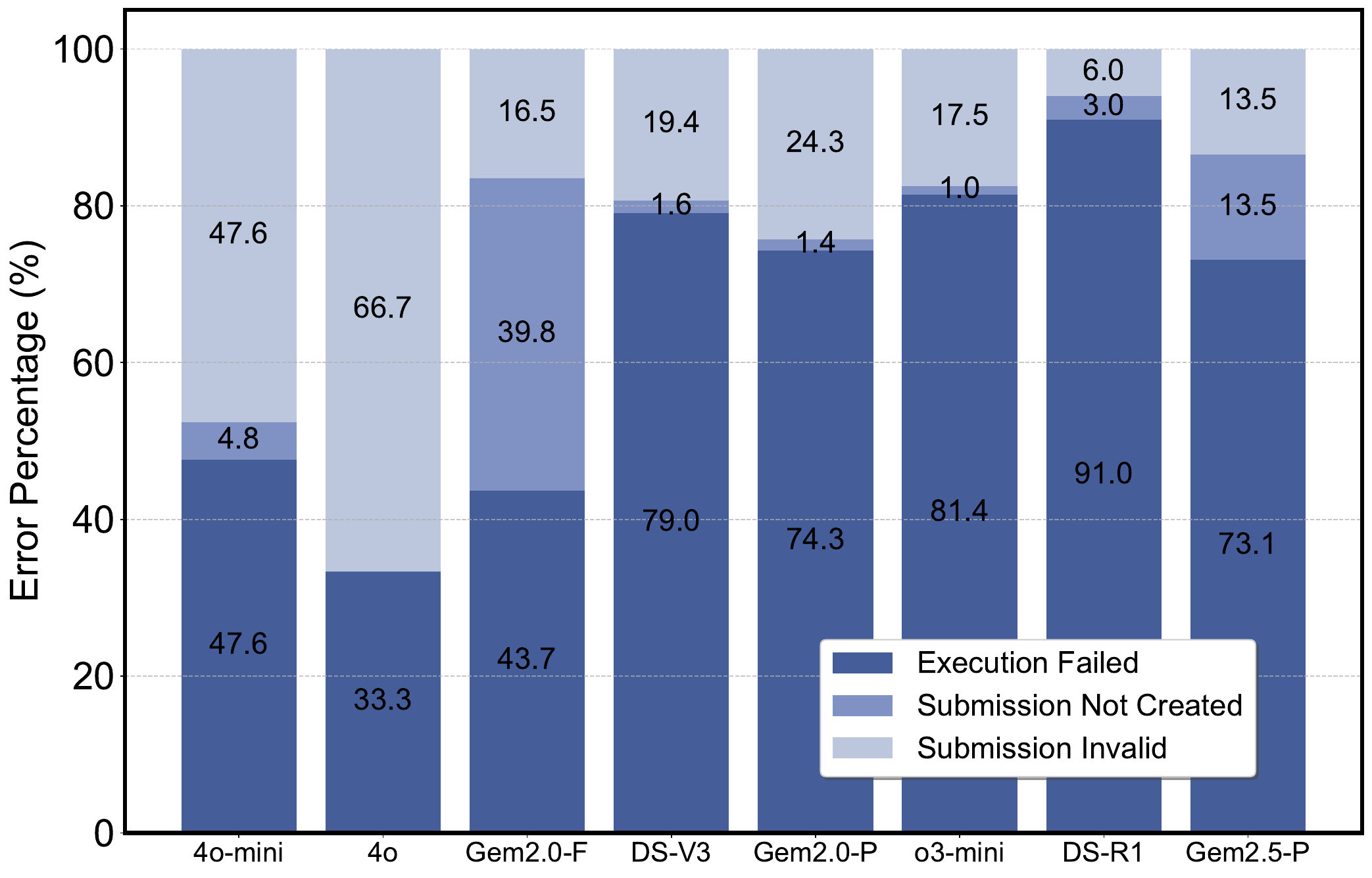}
  \caption{Execution error types.
  }
    \label{fig:error-types}
    % \vspace{-1ex}
\end{wrapfigure}
\paragraph{Execution Error Types.}
To further analyze the nature of execution errors, we categorized failures into "Execution Failed", "Submission Not Created", and "Submission Invalid" (Figure \ref{fig:error-types}). Stronger models, generating longer, more complex code, are more susceptible to "Execution Failed" errors. However, successful executions from these models frequently yield valid submissions and high performance. Models employing conservative validation strategies (e.g., \texttt{gpt-4o-mini}, \texttt{gpt-4o}) significantly reduce "Execution Failed" errors but continue to encounter issues related to submission creation and validity, indicating limitations in the validation approach.

\begin{wrapfigure}{r}{0.8\linewidth}
    \centering
    \vspace{-2ex}
    \includegraphics[width=\linewidth]{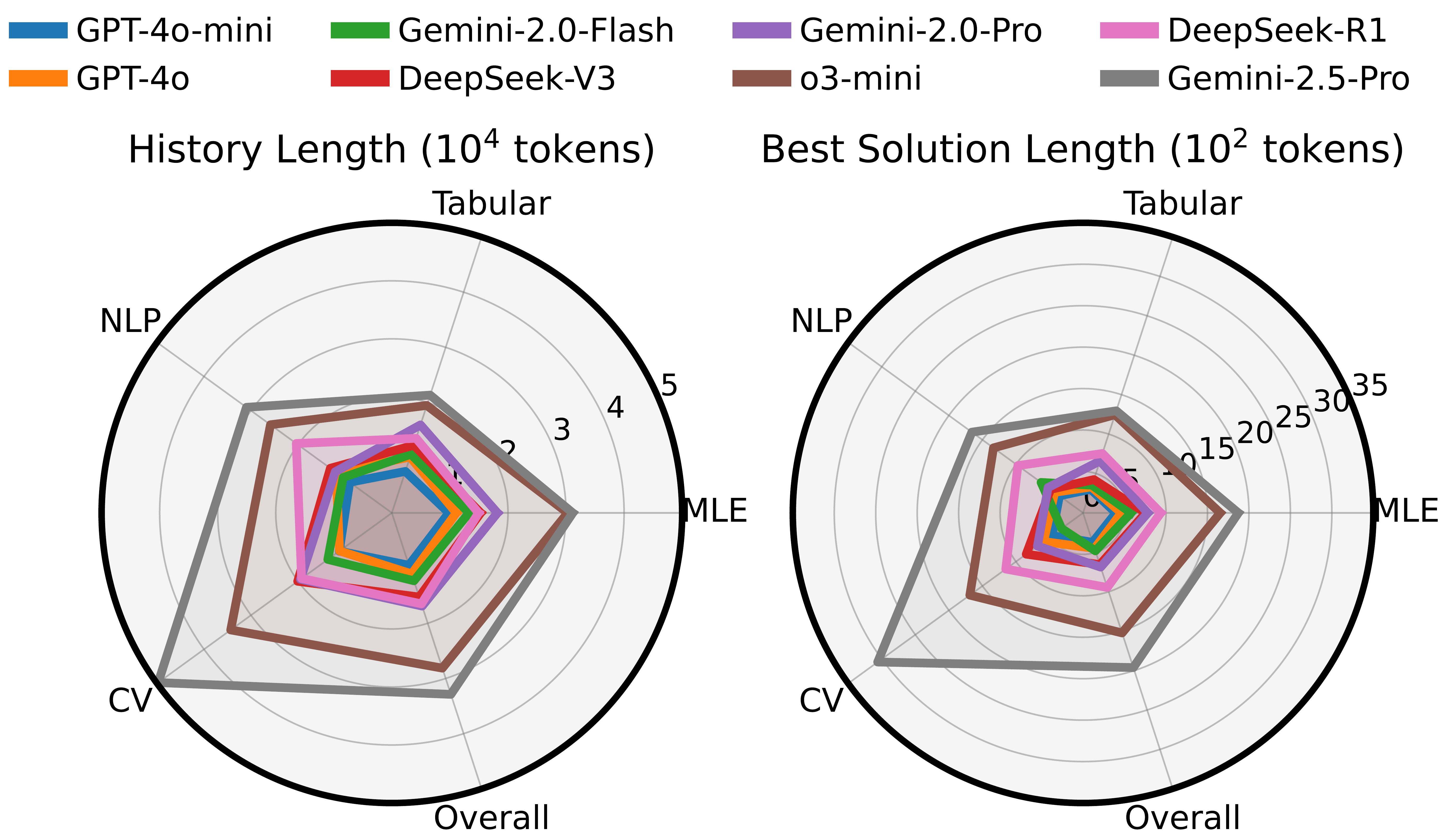}
    \caption{Total chat history length (left) and best solution length (right).}
    \label{fig:history-radar}
    \vspace{-1ex}
\end{wrapfigure}
\subsection{History Length and Solution Length}
Figure~\ref{fig:history-radar} shows both the total chat history length, including all interaction prompts, actions, and generated codes, and the length of the best solution generated by each model.
The total chat history length closely aligns with the best solution length, where both metrics positively correlate with overall model performance. Reasoning models typically generate notably longer solutions compared to non-reasoning models. Moreover, stronger-performing models frequently produce more extended solutions, which often correspond to higher performance scores. Although increased solution length does not inherently ensure superior outcomes, it generally indicates a model's capability to explore more intricate and sophisticated solution strategies, a hallmark predominantly observed in more capable reasoning models.

% \subsection{Prompting-based Agents}
% \todo{A huge table of prompt-based agents, covering different scafolds and LLMs}

% \subsection{Fine-Tuning Agents}
% \subsubsection{Data collection/Trajectory Sampling.}

% \todo{another table of base LLM with fine-tuned version}

% \subsection{Outcome Verifiers}
% \ycz{Reason: the interaction with MLE environments might take a lot of time (like code running time increase with the increasing size of training data and the complexity of methods), it is usually difficult to interact with the environment each time sampling data. 
% Training a verifier can help improve the agent performance and accelerate sampling process.}

\section{Conclusion}
\label{sec:conclusion}

We introduced \method, a Gym-style framework designed for training, evaluating, and benchmarking autonomous MLE agents. By providing an interactive and realistic environment built upon a large-scale collection of real-world Kaggle competitions, \method enables systematic experimentation, rigorous outcome verification, and iterative feedback crucial for advancing LLM-driven MLE workflows. Our extensive empirical evaluations established foundational baselines and highlighted both capabilities and critical limitations of current state-of-the-art models and agent architectures. We publicly release our framework, benchmarks, and leaderboard to encourage transparent comparison, reproducible research, and community-driven progress toward next-generation of fully autonomous MLE agents. Future work includes expanding \method to incorporate domain-specific deep research and support for multi-agent collaborative scenarios.

%%%%%%%%%%%%%%%%%%%%%%%%%%%%%%%%%%%%%%%%%%%%%%%%%%%%%%%%%%%%
\bibliographystyle{abbrv}
\bibliography{ref}

\newpage
\appendix
\section{Limitations and Broader Impacts}
\subsection{Limitations}

\noindent \textbf{Resource Limitations.} While \method provides a rich, interactive environment to train and evaluate LLM agents, it also imposes computational and storage demands. Executing full MLE pipelines-including data preprocessing, model training, and debugging-across hundreds of competition tasks requires potential API credits, CPU/GPU resources and large-scale disk storage. Although we provide competitions of varying difficulty levels and data sizes--organized in ascending order of size for user clarity--comprehensively training or evaluating MLE agents still demands substantial computational and storage resources.

\noindent \textbf{Data Privacy and Licensing.} \method is constructed atop real-world Kaggle competitions, each of which may be governed by different data usage licenses and privacy policies. Although the framework does not redistribute any proprietary data directly, it facilitates automated downloading and usage of datasets via the official Kaggle API. Users are solely responsible for reviewing, understanding, and adhering to the license agreements associated with each dataset. To assist in compliance, we provide direct links to the license terms and competition rules for all included tasks in \url{https://github.com/MLE-Dojo/MLE-Dojo/blob/main/prepare/licenses.json}.

\subsection{Broader Impacts}

\noindent \textbf{Potential Societal Impacts.} \method aims to advance the development of intelligent, autonomous agents capable of supporting and accelerating the machine learning engineering process. By enabling automation of repetitive and error-prone engineering tasks, the framework may help reduce the barrier to entry for ML practitioners, democratize access to model development tools, and empower domain experts (e.g., in healthcare, education, and scientific research) to more effectively apply machine learning solutions without requiring deep expertise in engineering workflows. Furthermore, the open-source nature of \method fosters reproducibility, transparency, and community-led innovation in the design and evaluation of LLM agents for real-world MLE tasks.

% \noindent \textbf{Potential Negative Societal Impacts.} 

\subsection{Ethical Statements}

\noindent \textbf{Data Usage and Consent.}  
\method does not host or redistribute any proprietary datasets. All tasks are built upon publicly available Kaggle competitions, and datasets are accessed via the official Kaggle API in compliance with their respective licenses. We ensure that no personal or sensitive information is retained within the framework. Users are explicitly informed that it is their responsibility to read and adhere to the terms of use and data handling policies provided by each competition. Where applicable, competitions selected for inclusion in the benchmark are reviewed to ensure they do not contain personally identifiable or ethically sensitive data.

\noindent \textbf{Fair Use and Responsible Deployment.}  
\method is intended solely for academic research, educational purposes, and the responsible development of AI systems. We discourage the use of this framework for automating competition participation without appropriate attribution or for circumventing fair use policies. We also emphasize the importance of transparent model development practices when using \method, particularly in scenarios where LLM agents are intended for downstream deployment or integration into critical systems.

\noindent \textbf{Bias and Representational Fairness.}  
While \method includes a broad spectrum of tasks across multiple ML domains, the selection of tasks is constrained by the availability of open competitions and may not fully represent all application domains or demographic contexts. Moreover, since LLM agents can inherit and amplify biases present in training data or model pretraining corpora, it is crucial that users critically evaluate outputs, particularly in sensitive contexts (e.g., healthcare, hiring, or education). Future versions of \method will explore the inclusion of bias detection and mitigation tools to support ethical agent development.

\noindent \textbf{Open-Source Commitment.}  
We commit to maintaining \method as an open-source project under a permissive license, enabling transparent inspection, community-driven development, and reproducibility. We welcome contributions that enhance the ethical robustness of the framework, including the addition of fairness metrics, safety guardrails, and improved documentation on responsible usage.

% We strictly follow data usage guidelines during the interaction with Microsoft Azure's Open AI API service. Although we utilize all publicly available datasets in this study, we have withdrawn from the human review process by completing and submitting the Azure OpenAI Additional Use Case Form to prevent any potential information leaks.
\section{Disclaimer}

The dataset is made available exclusively for educational and academic research purposes, with the intention of advancing scholarly investigations in relevant domains. Users of the dataset must adhere to the following terms and conditions:

\begin{itemize}[leftmargin=6mm]
    \item \textbf{Data Source and Accuracy}: Although reasonable efforts have been undertaken to curate and organize the dataset, the providers do not warrant the accuracy, completeness, or currency of the information. Users are strongly encouraged to independently verify the data and bear full responsibility for any analyses, interpretations, or conclusions derived from it.

    \item \textbf{Usage Restrictions}: This dataset is strictly limited to non-commercial use. Any commercial exploitation, product development, or profit-oriented application based on this dataset requires prior explicit written authorization from the dataset providers.

    \item \textbf{Privacy and Legal Compliance}: Users must ensure that their usage of the dataset complies with all relevant legal frameworks, particularly those concerning data privacy, protection, and security. The dataset providers shall not be held accountable for any legal liabilities or consequences resulting from improper or unauthorized usage.

    \item \textbf{Non-Infringement of Rights}: The dataset includes pre-processed content derived from external sources and is distributed solely for non-commercial research purposes. The dataset providers do not assert ownership over the original data and expressly acknowledge the rights of the original creators. It is the user's responsibility to ensure that their use of the dataset does not violate any copyright or intellectual property laws.

    \item \textbf{Disclaimer of Liability}: The dataset is provided "as is" without any express or implied warranties. The dataset providers shall not be liable for any direct, indirect, incidental, or consequential damages arising from the use of the dataset, including but not limited to financial loss, data misinterpretation, or third-party claims.
\end{itemize}

\section{Unified Data Structure}
\label{app:structure}

Users can flexibly incorporate new tasks originating from diverse sources, which can be standardized into a unified competition format with minimal overhead. These personalized competitions can then be seamlessly integrated into our environment for evaluation, benchmarking and training. 

\begin{figure}[h]
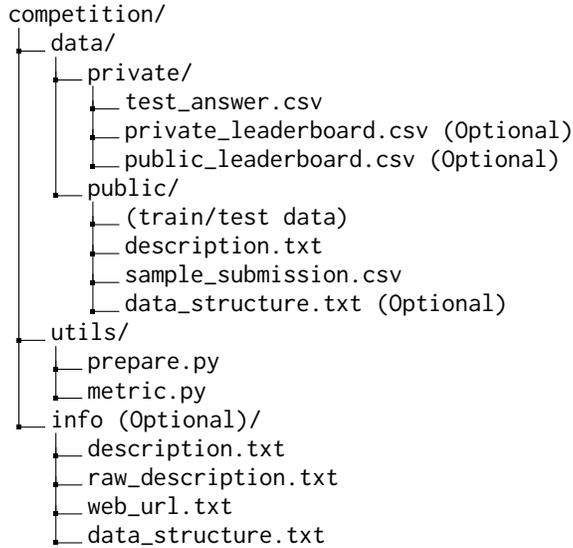


\dirtree{%
 .1 competition/.
 .2 data/.
 .3 private/.
 .4 test\_answer.csv.
 .4 private\_leaderboard.csv (Optional).
 .4 public\_leaderboard.csv (Optional).
 .3 public/.
 .4 (train/test data).
 .4 description.txt.
 .4 sample\_submission.csv.
 .4 data\_structure.txt (Optional).
 .2 utils/.
 .3 prepare.py.
 .3 metric.py.
 .2 info (Optional)/.
 .3 description.txt.
 .3 raw\_description.txt.
 .3 web\_url.txt.
 .3 data\_structure.txt.
}
\caption{Standardized directory structure for each competition.}
\label{fig:dir-structure}
\end{figure}

To ensure clarity, modularity, and scalability, each competition is organized under a unified and minimalistic directory structure. As shown in Figure~\ref{fig:dir-structure}, the root directory is composed of three major subdirectories: \texttt{data/}, \texttt{utils/}, and an optional \texttt{info/} folder. The \texttt{data/} directory is further divided into \texttt{private/} and \texttt{public/} subfolders, corresponding respectively to hidden and accessible phases of the competition. Each contains standardized files such as \texttt{test\_answer.csv}, \texttt{leaderboard.csv}, and input-output formats (e.g., \texttt{sample\_submission.csv}, \texttt{description.txt}), thereby facilitating consistent evaluation procedures. The \texttt{utils/} folder encapsulates task-specific scripts for data preparation and metric evaluation (i.e., \texttt{prepare.py}, \texttt{metric.py}). Lastly, the optional \texttt{info/} directory provides supplementary metadata such as web links, original descriptions, and data schema. This design adheres to a uniform organizational paradigm that simplifies integration, supports automation, and improves the transparency of competition configurations.

\section{Data Details}
\label{app:data}

To establish a unified and flexible framework capable of accommodating diverse and emerging tasks, we develop customized data preparation scripts for each competition. These scripts systematically structure datasets, generate representative example submissions, and enable efficient local testing and evaluation. Given that most Kaggle competitions lack publicly available labels for test datasets, our scripts carefully split the original training data into new, clearly defined training and test subsets. We further implement tailored evaluation metrics specific to each competition, derived from a common base metric class. 
We begin with an initial set of around 600 Kaggle competitions.
We've already manually reviewed them to confirm their suitability based on clarity of descriptions, data availability, and relevance of evaluation metrics, while excluding tasks with excessively large or unwieldy datasets. 
Our final collection balances challenge complexity with computational feasibility, yielding a practical, scalable, and user-friendly benchmark of 200 competitions as our first release. We plan progressive releases of processed datasets to support ongoing benchmarking efforts.

Table \ref{tab:kaggle-competitions} presents all competitions included in our first release, sorted in ascending order by data size. The associated tags provide information on each task's metric, category, modality, or domain. Users are encouraged to flexibly select tasks based on specific needs.

{
\begingroup % Keep font size change local
\small
% Use a smaller font size if the table is very wide/long

\begin{longtable}{>{\RaggedRight}p{3.5cm} l >{\RaggedRight}p{7.5cm}} % Adjust p{width} values as needed

% Caption and Label
\caption{Kaggle Competition Data Overview} \label{tab:kaggle-competitions}\\

% Table Header for the first page
\toprule
\textbf{Competition Name} & \textbf{Size} & \textbf{Tags} \\
\midrule
\endfirsthead

% Table Header for subsequent pages
\caption[]{Kaggle Competition Data Overview (continued)} \\ % Optional continued caption
\toprule
\textbf{Competition Name} & \textbf{Size} & \textbf{Tags} \\
\midrule
\endhead

% Footer for all pages except the last
\midrule
\multicolumn{3}{r}{\textit{Continued on next page}} \\
\endfoot

% Footer for the last page
\bottomrule
\endlastfoot

% --- Table Data Starts Here ---
playground-series-s3e12 & 25.56 kB & Beginner, Tabular, Binary Classification, Custom Metric \\
titanic & 93.08 kB & Binary Classification, Tabular, Beginner, Categorization Accuracy \\
playground-series-s3e5 & 225.6 kB & Beginner, Tabular, Cohen Kappa Score \\
playground-series-s3e13 & 284.43 kB & Beginner, Tabular, Multiclass Classification, Health Conditions, MAP\@\{K\} \\
mercedes-benz-greener-manufacturing & 351.45 kB & Automobiles and Vehicles, Regression, Tabular, $R^2$ score (coefficient of determination) \\
icr-identify-age-related-conditions & 356.88 kB & Tabular, Binary Classification, Health, Weighted Multiclass Loss \\
kaggle-llm-science-exam & 364.21 kB & Physics, NLP, MAP\@\{K\} \\
playground-series-s3e22 & 386.6 kB & Beginner, Tabular, Multiclass Classification, Animals, Health, F1 Score \\
playground-series-s3e3 & 455.63 kB & Tabular, Beginner, Binary Classification, Business, Area Under Receiver Operating Characteristic Curve \\
playground-series-s3e9 & 485.23 kB & Beginner, Tabular, Regression, Root Mean Squared Error \\
kobe-bryant-shot-selection & 708.3 kB & Basketball, Binary Classification, Tabular, Log Loss \\
tabular-playground-series-jul-2021 & 826.96 kB & Tabular, Pollution, Time Series Analysis, Mean Columnwise Root Mean Squared Logarithmic Error \\
bike-sharing-demand & 1.12 MB & Cycling, Tabular, Time Series Analysis, Root Mean Squared Logarithmic Error \\
home-data-for-ml-course & 1.15 MB & Mean Absolute Error \\
spaceship-titanic & 1.24 MB & Beginner, Tabular, Binary Classification, Categorization Accuracy \\
covid19-global-forecasting-week-2 & 1.26 MB & Coronavirus, Tabular, Mean Columnwise Root Mean Squared Logarithmic Error \\
covid19-global-forecasting-week-3 & 1.41 MB & Tabular, Coronavirus, Mean Columnwise Root Mean Squared Logarithmic Error \\
nlp-getting-started & 1.43 MB & Text, Binary Classification, NLP, Custom Metric \\
covid19-global-forecasting-week-1 & 1.63 MB & Coronavirus, Tabular, Mean Columnwise Root Mean Squared Logarithmic Error \\
tabular-playground-series-jan-2022 & 1.73 MB & Tabular, Time Series Analysis, SMAPE \\
playground-series-s3e2 & 1.88 MB & Binary Classification, Tabular, Health Conditions, Beginner, Area Under Receiver Operating Characteristic Curve \\
spooky-author-identification & 1.9 MB & Multiclass Classification, Literature, Linguistics, Multiclass Loss \\
covid19-global-forecasting-week-4 & 1.95 MB & Tabular, Coronavirus, Mean Columnwise Root Mean Squared Logarithmic Error \\
liberty-mutual-group-property-inspection-prediction & 2 MB & Housing, Normalized Gini Index \\
playground-series-s3e25 & 2.11 MB & Beginner, Tabular, Regression, Earth Science, Median Absolute Error \\
us-patent-phrase-to-phrase-matching & 2.14 MB & NLP, Text, PearsonCorrelationCoefficient \\
movie-review-sentiment-analysis-kernels-only & 2.44 MB & Text, Multiclass Classification, Categorization Accuracy \\
playground-series-s3e6 & 2.7 MB & Beginner, Tabular, Housing, Root Mean Squared Error \\
commonlitreadabilityprize & 2.93 MB & Text, Regression, Root Mean Squared Error \\
playground-series-s3e14 & 2.99 MB & Beginner, Tabular, Regression, Mean Absolute Error \\
detecting-insults-in-social-commentary & 3.02 MB & Area Under Receiver Operating Characteristic Curve \\
walmart-recruiting-store-sales-forecasting & 3.22 MB & Time Series Analysis, Weighted Mean Absolute Error \\
prudential-life-insurance-assessment & 3.4 MB & Tabular, QuadraticWeightedKappa \\
tweet-sentiment-extraction & 3.86 MB & Text, Internet, Custom Metric \\
playground-series-s3e7 & 3.86 MB & Beginner, Tabular, Binary Classification, Area Under Receiver Operating Characteristic Curve \\
llm-detect-ai-generated-text & 4.43 MB & Education, Primary and Secondary Schools, Binary Classification, Text Generation, Roc Auc Score \\
unimelb & 4.53 MB & Area Under Receiver Operating Characteristic Curve \\
playground-series-s4e2 & 4.59 MB & Beginner, Time Series Analysis, Tabular, Multiclass Classification, Accuracy Score \\
playground-series-s4e3 & 5.49 MB & Beginner, Tabular, Multiclass Classification, Binary Classification, Manufacturing, Mean Columnwise Area Under Receiver Operating Characteristic Curve \\
tabular-playground-series-sep-2022 & 5.73 MB & Tabular, SMAPE \\
nomad2018-predict-transparent-conductors & 6.24 MB & Chemistry, Mean Columnwise Root Mean Squared Logarithmic Error \\
amazon-employee-access-challenge & 6.39 MB & Area Under Receiver Operating Characteristic Curve \\
playground-series-s3e1 & 6.48 MB & Regression, Tabular, Housing, Beginner, Root Mean Squared Error \\
tabular-playground-series-aug-2022 & 7.21 MB & Tabular, Binary Classification, Area Under Receiver Operating Characteristic Curve \\
playground-series-s3e18 & 7.63 MB & Beginner, Tabular, Binary Classification, Multilabel Classification, Roc Auc Score \\
poker-rule-induction & 8.3 MB & Multiclass Classification, Card Games, Tabular, Categorization Accuracy \\
playground-series-s4e4 & 8.4 MB & Beginner, Tabular, Regression, Mean Squared Log Error \\
allstate-purchase-prediction-challenge & 8.97 MB & Categorization Accuracy \\
playground-series-s3e16 & 9.05 MB & Beginner, Tabular, Regression, Animals, Mean Absolute Error \\
feedback-prize-english-language-learning & 9.3 MB & NLP, Education, Primary and Secondary Schools, Custom Metric \\
dont-call-me-turkey & 11.04 MB & Binary Classification, Tabular, Animals, Area Under Receiver Operating Characteristic Curve \\
tabular-playground-series-apr-2021 & 12.66 MB & Beginner, Tabular, Binary Classification, Categorization Accuracy \\
playground-series-s3e19 & 12.79 MB & Beginner, Tabular, Time Series Analysis, SMAPE \\
playground-series-s3e17 & 12.86 MB & Beginner, Tabular, Roc Auc Score \\
GiveMeSomeCredit & 14.47 MB & Area Under Receiver Operating Characteristic Curve \\
google-quest-challenge & 14.85 MB & Text, NLP, Mean Columnwise Spearman's r (rank correlation coefficient) \\
forest-cover-type-kernels-only & 15.38 MB & Tabular, Forestry, Categorization Accuracy \\
playground-series-s4e6 & 16.2 MB & Beginner, Tabular, Education, Accuracy Score \\
novozymes-enzyme-stability-prediction & 16.39 MB & Chemistry, SpearmanR \\
integer-sequence-learning & 17.91 MB & Tabular, Categorization Accuracy \\
random-acts-of-pizza & 17.97 MB & Binary Classification, Text, Internet, Area Under Receiver Operating Characteristic Curve \\
playground-series-s3e23 & 18.63 MB & Beginner, Tabular, Binary Classification, Roc Auc Score \\
demand-forecasting-kernels-only & 18.7 MB & Tabular, SMAPE \\
playground-series-s3e8 & 18.96 MB & Beginner, Tabular, Regression, Root Mean Squared Error \\
playground-series-s3e10 & 20.94 MB & Beginner, Tabular, Log Loss \\
playground-series-s4e1 & 21.65 MB & Beginner, Tabular, Binary Classification, Banking, Roc Auc Score \\
afsis-soil-properties & 21.7 MB & Mean Columnwise Root Mean Squared Error \\
sberbank-russian-housing-market & 22.71 MB & Banking, Housing, Regression, Tabular, Root Mean Squared Logarithmic Error \\
playground-series-s3e24 & 22.79 MB & Beginner, Tabular, Binary Classification, Health, Roc Auc Score \\
aerial-cactus-identification & 25.4 MB & Earth and Nature, Image, Plants, Area Under Receiver Operating Characteristic Curve \\
crowdflower-weather-twitter & 25.4 MB & Root Mean Squared Error \\
predicting-red-hat-business-value & 26.74 MB & Tabular, Business, Area Under Receiver Operating Characteristic Curve \\
DontGetKicked & 29.52 MB & Gini Index \\
battlefin-s-big-data-combine-forecasting-challenge & 30.23 MB & Mean Absolute Error \\
bioresponse & 31.06 MB & Log Loss \\
tabular-playground-series-mar-2022 & 31.4 MB & Tabular, Time Series Analysis, Cities and Urban Areas, Regression, Mean Absolute Error \\
chaii-hindi-and-tamil-question-answering & 31.8 MB & Text, Languages, Custom Metric \\
nbme-score-clinical-patient-notes & 35.73 MB & Text, Medicine, Education, NLP, Custom Metric \\
leaf-classification & 36.05 MB & Image, Multiclass Classification, Multiclass Loss \\
learning-agency-lab-automated-essay-scoring-2 & 36.2 MB & Education, NLP, Primary and Secondary Schools, Cohen Kappa Score \\
20-newsgroups-ciphertext-challenge & 36.97 MB & Multiclass Classification, Text, F-Score (Macro) \\
dont-overfit-ii & 38.6 MB & Tabular, Binary Classification, Area Under Receiver Operating Characteristic Curve \\
rossmann-store-sales & 39.85 MB & Tabular, Time Series Analysis, Root Mean Square Percentage Error \\
sf-crime & 45.24 MB & Multiclass Classification, Tabular, Crime, Multiclass Loss \\
imaterialist-challenge-furniture-2018 & 48.59 MB & MeanBestErrorAtK \\
playground-series-s3e11 & 49.83 MB & Beginner, Tabular, Regression, Mean Squared Log Error \\
word2vec-nlp-tutorial & 54.37 MB & Text, Binary Classification, Movies and TV Shows, Area Under Receiver Operating Characteristic Curve \\
jigsaw-toxic-comment-classification-challenge & 55.18 MB & Text, Mean Columnwise Area Under Receiver Operating Characteristic Curve \\
see-click-predict-fix & 55.75 MB & Root Mean Squared Logarithmic Error \\
denoising-dirty-documents & 58.81 MB & Image, Root Mean Squared Error \\
homesite-quote-conversion & 65.13 MB & Binary Classification, Tabular, Area Under Receiver Operating Characteristic Curve \\
airbnb-recruiting-new-user-bookings & 67.85 MB & Hotels and Accommodations, Recommender Systems, Tabular, NDCG\@\{K\} \\
cat-in-the-dat & 67.96 MB & Tabular, Binary Classification, Area Under Receiver Operating Characteristic Curve \\
tmdb-box-office-prediction & 70.24 MB & Tabular, Movies and TV Shows, Root Mean Squared Logarithmic Error \\
home-depot-product-search-relevance & 72.93 MB & Tabular, Root Mean Squared Error \\
facial-keypoints-detection & 80.86 MB & Image, Root Mean Squared Error \\
nyc-taxi-trip-duration & 89.91 MB & Tabular, Regression, Root Mean Squared Logarithmic Error \\
text-normalization-challenge-english-language & 95.51 MB & Linguistics, Languages, Text, Categorization Accuracy \\
bnp-paribas-cardif-claims-management & 103.75 MB & Banking, Tabular, Binary Classification, Log Loss \\
playground-series-s4e5 & 104.17 MB & Beginner, Tabular, Logistic Regression, R2 Score \\
wsdm-cup-multilingual-chatbot-arena & 113.98 MB & Languages, Text Conversation, Accuracy Score \\
santander-customer-satisfaction & 119.04 MB & Tabular, Binary Classification, Banking, Area Under Receiver Operating Characteristic Curve \\
playground-series-s3e20 & 119.66 MB & Beginner, Tabular, Time Series Analysis, Root Mean Squared Error \\
text-normalization-challenge-russian-language & 125.87 MB & Text, Linguistics, Languages, Categorization Accuracy \\
tabular-playground-series-mar-2021 & 131.65 MB & Tabular, Logistic Regression, Binary Classification, Area Under Receiver Operating Characteristic Curve \\
allstate-claims-severity & 142.87 MB & Regression, Tabular, Mean Absolute Error \\
tabular-playground-series-jan-2021 & 144.43 MB & Tabular, Regression, Root Mean Squared Error \\
cat-in-the-dat-ii & 145.84 MB & Binary Classification, Area Under Receiver Operating Characteristic Curve \\
liverpool-ion-switching & 146.08 MB & Biology, F-Score (Macro) \\
tabular-playground-series-feb-2021 & 154.66 MB & Regression, Tabular, Root Mean Squared Error \\
yelp-recsys-2013 & 179.46 MB & Root Mean Squared Error \\
lmsys-chatbot-arena & 184.19 MB & Languages, Text Conversation, Log Loss \\
llm-classification-finetuning & 184.19 MB & Languages, Text Conversation, Log Loss \\
stumbleupon & 196.18 MB & Text, Tabular, Internet, Area Under Receiver Operating Characteristic Curve \\
playground-series-s3e4 & 198.09 MB & Tabular, Binary Classification, Beginner, Area Under Receiver Operating Characteristic Curve \\
the-winton-stock-market-challenge & 213.13 MB & Finance, Tabular, Weighted Mean Absolute Error \\
conways-reverse-game-of-life-2020 & 251.11 MB & Simulations, Board Games, Custom Metric \\
facebook-recruiting-iv-human-or-bot & 260.68 MB & Binary Classification, Tabular, Internet, Area Under Receiver Operating Characteristic Curve \\
the-icml-2013-whale-challenge-right-whale-redux & 293.14 MB & Area Under Receiver Operating Characteristic Curve \\
porto-seguro-safe-driver-prediction & 300.58 MB & Tabular, Binary Classification, Normalized Gini Index \\
statoil-iceberg-classifier-challenge & 302.1 MB & Binary Classification, Weather and Climate, Image, Log Loss \\
tabular-playground-series-aug-2021 & 337.38 MB & Tabular, Regression, Banking, Root Mean Squared Error \\
flavours-of-physics-kernels-only & 436.14 MB & Custom Metric \\
tgs-salt-identification-challenge & 483.07 MB & Geology, Image, Custom Metric \\
linking-writing-processes-to-writing-quality & 485.71 MB & Education, NLP, Primary and Secondary Schools, Mean Squared Error \\
grupo-bimbo-inventory-demand & 502.44 MB & Tabular, Food, Root Mean Squared Logarithmic Error \\
quora-question-pairs & 523.24 MB & Text, Tabular, Linguistics, Internet, Log Loss \\
tabular-playground-series-may-2022 & 574.22 MB & Tabular, Binary Classification, Area Under Receiver Operating Characteristic Curve \\
bigquery-geotab-intersection-congestion & 577.97 MB & Tabular, Regression, Cities and Urban Areas, Geospatial Analysis, Root Mean Squared Error \\
mlsp-2013-birds & 585.1 MB & Area Under Receiver Operating Characteristic Curve \\
santander-customer-transaction-prediction & 606.35 MB & Banking, Tabular, Binary Classification, Area Under Receiver Operating Characteristic Curve \\
loan-default-prediction & 611.66 MB & Mean Absolute Error \\
global-wheat-detection & 643.57 MB & Image, Plants, Custom Metric \\
tabular-playground-series-dec-2021 & 693.17 MB & Tabular, Multiclass Classification, Categorization Accuracy \\
ventilator-pressure-prediction & 698.79 MB & Tabular, Medicine, Biology, Mean Absolute Error \\
whale-categorization-playground & 726.74 MB & Image, Animals, MAP\@\{K\} \\
dog-breed-identification & 750.43 MB & Multiclass Classification, Animals, Image, Multiclass Loss \\
plant-pathology-2020-fgvc7 & 823.79 MB & Image, Agriculture, Mean Columnwise Area Under Receiver Operating Characteristic Curve \\
dogs-vs-cats-redux-kernels-edition & 854.51 MB & Image, Animals, Binary Classification, Log Loss \\
benchmark-bond-trade-price-challenge & 910.54 MB & Weighted Mean Absolute Error \\
pubg-finish-placement-prediction & 965.66 MB & Video Games, Tabular, Mean Absolute Error \\
instant-gratification & 972.6 MB & Tabular, Binary Classification, Area Under Receiver Operating Characteristic Curve \\
tabular-playground-series-nov-2021 & 1.04 GB & Tabular, Binary Classification, Area Under Receiver Operating Characteristic Curve \\
petfinder-pawpularity-score & 1.04 GB & Image, Root Mean Squared Error \\
santander-value-prediction-challenge & 1.08 GB & Banking, Finance, Root Mean Squared Logarithmic Error \\
champs-scalar-coupling & 1.22 GB & Chemistry, Tabular, Regression, Custom Metric \\
avazu-ctr-prediction & 1.28 GB & Log Loss \\
tabular-playground-series-sep-2021 & 1.37 GB & Tabular, Binary Classification, Area Under Receiver Operating Characteristic Curve \\
billion-word-imputation & 1.7 GB & Text, Linguistics, Levenshtein Mean \\
plant-seedlings-classification & 1.81 GB & Plants, Image, Multiclass Classification, Custom Metric \\
tabular-playground-series-feb-2022 & 1.87 GB & Tabular, Multiclass Classification, Categorization Accuracy \\
shopee-product-matching & 1.92 GB & Image, Text, Retail and Shopping, Custom Metric \\
AI4Code & 2.16 GB & NLP, Text, Computer Science, Custom Metric \\
the-nature-conservancy-fisheries-monitoring & 2.27 GB & Image, Multiclass Classification, Multiclass Loss \\
jigsaw-unintended-bias-in-toxicity-classification & 2.38 GB & Text, NLP, Custom Metric \\
uw-madison-gi-tract-image-segmentation & 2.47 GB & Image, Medicine, Dice3DHausdorff \\
ashrae-energy-prediction & 2.61 GB & Tabular, Energy, Root Mean Squared Logarithmic Error \\
stanford-covid-vaccine & 2.68 GB & Biology, Biotechnology, Coronavirus, Public Health, Custom Metric \\
acquire-valued-shoppers-challenge & 3.07 GB & Area Under Receiver Operating Characteristic Curve \\
facebook-recruiting-iii-keyword-extraction & 3.19 GB & Custom Metric \\
invasive-species-monitoring & 3.35 GB & Image, Plants, Area Under Receiver Operating Characteristic Curve \\
tabular-playground-series-oct-2021 & 3.49 GB & Tabular, Binary Classification, Area Under Receiver Operating Characteristic Curve \\
kuzushiji-recognition & 4.51 GB & Image, Multiclass Classification, History, Japan, Custom Metric \\
nfl-player-contact-detection & 5.01 GB & Health, Football, Video, Tabular, Matthews correlation coefficient \\
bengaliai-cv19 & 5.18 GB & Image, Multiclass Classification, Weighted Categorization Accuracy \\
new-york-city-taxi-fare-prediction & 5.7 GB & Regression, Tabular, Root Mean Squared Error \\
cassava-leaf-disease-classification & 6.19 GB & Image, Multiclass Classification, Plants, Categorization Accuracy \\
quora-insincere-questions-classification & 6.56 GB & Text, Binary Classification, Custom Metric \\
histopathologic-cancer-detection & 7.76 GB & Cancer, Medicine, Research, Area Under Receiver Operating Characteristic Curve \\
microsoft-malware-prediction & 8.47 GB & Area Under Receiver Operating Characteristic Curve \\
bms-molecular-translation & 8.87 GB & Chemistry, Image, Levenshtein Mean \\
aptos2019-blindness-detection & 10.22 GB & Image, Multiclass Classification, Medicine, Healthcare, QuadraticWeightedKappa \\
ranzcr-clip-catheter-line-classification & 13.13 GB & Image, Multilabel Classification, Mean Columnwise Area Under Receiver Operating Characteristic Curve \\
plant-pathology-2021-fgvc8 & 16.1 GB & Image, Plants, Custom Metric \\
smartphone-decimeter-2022 & 22.9 GB & Geospatial Analysis, Signal Processing, Research, Tabular, Mobile and Wireless, Custom Metric \\
multi-modal-gesture-recognition & 23.23 GB & Custom Metric \\
osic-pulmonary-fibrosis-progression & 23.99 GB & Image, Healthcare, Laplace Log Likelihood \\
freesound-audio-tagging-2019 & 26.15 GB & Audio, Weighted Label Ranking Average Precision \\
hms-harmful-brain-activity-classification & 26.4 GB & Tabular, Research, Signal Processing, Healthcare, Kullback Leibler Divergence \\
hotel-id-2021-fgvc8 & 26.65 GB & Image, Public Safety, MAP\@\{K\} \\
imet-2020-fgvc7 & 29.46 GB & F-Score Beta (Micro) \\
predict-volcanic-eruptions-ingv-oe & 31.25 GB & Signal Processing, Geology, Physics, Mean Absolute Error \\
airbus-ship-detection & 31.41 GB & Image, IntersectionOverUnionObjectSegmentationBeta \\
alaska2-image-steganalysis & 32.27 GB & Custom Metric \\
hubmap-kidney-segmentation & 32.97 GB & Image, Health, Biology, Dice \\
h-and-m-personalized-fashion-recommendations & 34.56 GB & Recommender Systems, Retail and Shopping, MAP\@\{K\} \\
draper-satellite-image-chronology & 36.07 GB & Image, MASpearmanR \\
vesuvius-challenge-ink-detection & 37.02 GB & Image Segmentation, History, Image Text Recognition, DiceFBeta \\
iwildcam-2019-fgvc6 & 46.68 GB & Image, Multiclass Classification, F-Score (Macro) \\
herbarium-2020-fgvc7 & 63.16 GB & Image, Plants, F-Score (Macro) \\
cdiscount-image-classification-challenge & 78.12 GB & Multiclass Classification, Categorization Accuracy \\
inaturalist-2019-fgvc6 & 87.76 GB & MeanBestErrorAtK \\
icecube-neutrinos-in-deep-ice & 117.18 GB & Tabular, Astronomy, Physics, MeanAngularError \\
iwildcam-2020-fgvc7 & 118.59 GB & Multiclass Classification, Biology, Categorization Accuracy \\
siim-covid19-detection & 128.51 GB & Image, Multilabel Classification, Custom Metric \\
rsna-miccai-brain-tumor-radiogenomic-classification & 136.85 GB & Image, Binary Classification, Healthcare, Area Under Receiver Operating Characteristic Curve \\
seti-breakthrough-listen & 156.02 GB & Astronomy, Signal Processing, Science and Technology, Area Under Receiver Operating Characteristic Curve \\
herbarium-2021-fgvc8 & 161.9 GB & Image, Plants, F-Score (Macro) \\
herbarium-2022-fgvc9 & 163.17 GB & Plants, Image, F-Score (Macro) \\
vinbigdata-chest-xray-abnormalities-detection & 205.96 GB & Image, Healthcare, Custom Metric \\
% --- Table Data Ends Here ---

\end{longtable}
\endgroup % End local font size change  

}

Table \ref{tab:kaggle-eval} presents 50 competitions used for evaluation in our benchmark, encompassing four categories of tasks: MLE-Lite, Tabular, NLP, and CV.
% Optional: Adjust font size for the whole table if needed
% {\small % or \footnotesize
{
\small
\begin{longtable}{>{\RaggedRight}p{3.5cm} l >{\RaggedRight}p{7.5cm}} % Adjust p{width} values as needed

  %--- Caption and Headers ---
  \caption{Competitions for Evaluation} \label{tab:kaggle-eval} \\
  \toprule
  \textbf{Competition Name} & \textbf{Size} & \textbf{Tags} \\
  \midrule
  \endfirsthead % Header for the first page

  % \multicolumn{3}{@{}l}{\tablename\ \thetable{} -- Continued from previous page} \\ % Optional: Add table number continuation text
  \toprule
  \textbf{Competition Name} & \textbf{Size} & \textbf{Tags} \\
  \midrule
  \endhead % Header for subsequent pages

  %--- Footers ---
  \midrule
  \multicolumn{3}{r@{}}{\textit{Continued on next page}} \\
  \endfoot % Footer for pages before the last

  \bottomrule
  \endlastfoot % Footer for the very last page

  %--- Table Body ---
  % --- MLE-Lite Competitions ---
  \multicolumn{3}{c}{\textit{MLE-Lite Competitions}} \\ \midrule
  spooky-author-identification & 1.9 MB & \RaggedRight Multiclass Classification, Literature, Linguistics, Multiclass Loss \\ % Use \RaggedRight for p columns
  detecting-insults-in-social-commentary & 3.02 MB & \RaggedRight Area Under Receiver Operating Characteristic Curve \\
  nomad2018-predict-transparent-conductors & 6.24 MB & \RaggedRight Chemistry, Mean Columnwise Root Mean Squared Logarithmic Error \\
  random-acts-of-pizza & 17.97 MB & \RaggedRight Binary Classification, Text, Internet, Area Under Receiver Operating Characteristic Curve \\
  aerial-cactus-identification & 25.4 MB & \RaggedRight Earth and Nature, Image, Plants, Area Under Receiver Operating Characteristic Curve \\
  leaf-classification & 36.05 MB & \RaggedRight Image, Multiclass Classification, Multiclass Loss \\
  jigsaw-toxic-comment-classification-challenge & 55.18 MB & \RaggedRight Text, Mean Columnwise Area Under Receiver Operating Characteristic Curve \\
  denoising-dirty-documents & 58.81 MB & \RaggedRight Image, Root Mean Squared Error \\
  text-normalization-challenge-english-language & 95.51 MB & \RaggedRight Linguistics, Languages, Text, Categorization Accuracy \\
  text-normalization-challenge-russian-language & 125.87 MB & \RaggedRight Text, Linguistics, Languages, Categorization Accuracy \\
  the-icml-2013-whale-challenge-right-whale-redux & 293.14 MB & \RaggedRight Area Under Receiver Operating Characteristic Curve \\
  tabular-playground-series-may-2022 & 574.22 MB & \RaggedRight Tabular, Binary Classification, Area Under Receiver Operating Characteristic Curve \\
  mlsp-2013-birds & 585.1 MB & \RaggedRight Area Under Receiver Operating Characteristic Curve \\
  tabular-playground-series-dec-2021 & 693.17 MB & \RaggedRight Tabular, Multiclass Classification, Categorization Accuracy \\
  dog-breed-identification & 750.43 MB & \RaggedRight Multiclass Classification, Animals, Image, Multiclass Loss \\
  plant-pathology-2020-fgvc7 & 823.79 MB & \RaggedRight Image, Agriculture, Mean Columnwise Area Under Receiver Operating Characteristic Curve \\
  dogs-vs-cats-redux-kernels-edition & 854.51 MB & \RaggedRight Image, Animals, Binary Classification, Log Loss \\
  new-york-city-taxi-fare-prediction & 5.7 GB & \RaggedRight Regression, Tabular, Root Mean Squared Error \\
  histopathologic-cancer-detection & 7.76 GB & \RaggedRight Cancer, Medicine, Research, Area Under Receiver Operating Characteristic Curve \\
  aptos2019-blindness-detection & 10.22 GB & \RaggedRight Image, Multiclass Classification, Medicine, Healthcare, QuadraticWeightedKappa \\
  ranzcr-clip-catheter-line-classification & 13.13 GB & \RaggedRight Image, Multilabel Classification, Mean Columnwise Area Under Receiver Operating Characteristic Curve \\
  \midrule

  % --- NLP Competitions ---
  \multicolumn{3}{c}{\textit{NLP Competitions}} \\ \midrule
  kaggle-llm-science-exam & 364.21 kB & \RaggedRight Physics, NLP, MAP@{K} \\
  llm-detect-ai-generated-text & 4.43 MB & \RaggedRight Education, Primary and Secondary Schools, Binary Classification, Text Generation, Roc Auc Score \\
  20-newsgroups-ciphertext-challenge & 36.97 MB & \RaggedRight Multiclass Classification, Text, F-Score (Macro) \\
  lmsys-chatbot-arena & 184.19 MB & \RaggedRight Languages, Text Conversation, Log Loss \\
  stumbleupon & 196.18 MB & \RaggedRight Text, Tabular, Internet, Area Under Receiver Operating Characteristic Curve \\
  linking-writing-processes-to-writing-quality & 485.71 MB & \RaggedRight Education, NLP, Primary and Secondary Schools, Mean Squared Error \\
  quora-question-pairs & 523.24 MB & \RaggedRight Text, Tabular, Linguistics, Internet, Log Loss \\
  AI4Code & 2.16 GB & \RaggedRight NLP, Text, Computer Science, Custom Metric \\
  quora-insincere-questions-classification & 6.56 GB & \RaggedRight Text, Binary Classification, Custom Metric \\
  \midrule

  % --- CV Competitions ---
  \multicolumn{3}{c}{\textit{CV Competitions}} \\ \midrule
  facial-keypoints-detection & 80.86 MB & \RaggedRight Image, Root Mean Squared Error \\
  whale-categorization-playground & 726.74 MB & \RaggedRight Image, Animals, MAP@{K} \\
  petfinder-pawpularity-score & 1.04 GB & \RaggedRight Image, Root Mean Squared Error \\
  bengaliai-cv19 & 5.18 GB & \RaggedRight Image, Multiclass Classification, Weighted Categorization Accuracy \\
  cassava-leaf-disease-classification & 6.19 GB & \RaggedRight Image, Multiclass Classification, Plants, Categorization Accuracy \\
  bms-molecular-translation & 8.87 GB & \RaggedRight Chemistry, Image, Levenshtein Mean \\
  imet-2020-fgvc7 & 29.46 GB & \RaggedRight F-Score Beta (Micro) \\
  airbus-ship-detection & 31.41 GB & \RaggedRight Image, IntersectionOverUnionObjectSegmentationBeta \\
  alaska2-image-steganalysis & 32.27 GB & \RaggedRight Custom Metric \\
  draper-satellite-image-chronology & 36.07 GB & \RaggedRight Image, MASpearmanR \\
  \midrule

  % --- Tabular Competitions ---
  \multicolumn{3}{c}{\textit{Tabular Competitions}} \\ \midrule
  demand-forecasting-kernels-only & 18.7 MB & \RaggedRight Tabular, SMAPE \\
  dont-overfit-ii & 38.6 MB & \RaggedRight Tabular, Binary Classification, Area Under Receiver Operating Characteristic Curve \\
  santander-customer-satisfaction & 119.04 MB & \RaggedRight Tabular, Binary Classification, Banking, Area Under Receiver Operating Characteristic Curve \\
  liverpool-ion-switching & 146.08 MB & \RaggedRight Biology, F-Score (Macro) \\
  conways-reverse-game-of-life-2020 & 251.11 MB & \RaggedRight Simulations, Board Games, Custom Metric \\
  porto-seguro-safe-driver-prediction & 300.58 MB & \RaggedRight Tabular, Binary Classification, Normalized Gini Index \\
  santander-customer-transaction-prediction & 606.35 MB & \RaggedRight Banking, Tabular, Binary Classification, Area Under Receiver Operating Characteristic Curve \\
  ventilator-pressure-prediction & 698.79 MB & \RaggedRight Tabular, Medicine, Biology, Mean Absolute Error \\
  instant-gratification & 972.6 MB & \RaggedRight Tabular, Binary Classification, Area Under Receiver Operating Characteristic Curve \\
  santander-value-prediction-challenge & 1.08 GB & \RaggedRight Banking, Finance, Root Mean Squared Logarithmic Error \\

\end{longtable}
% } % End of \small or \footnotesize block if used
}

We ranked the difficulty of competitions based on the average HumanRank scores achieved by all models across each competition. Table~\ref{tab:difficulty} presents all competitions for evaluation ranked by difficulty from easiest to hardest. This difficulty ranking provides a partial indication of which competitions MLE Agents perform relatively better in comparison to human competitors. However, due to potential biases introduced by factors such as dataset splits, the ranking should be interpreted as a reference rather than a definitive measure.
{
\begin{center}\label{tab:difficulty}
\small
\begin{longtable}{@{}p{6cm} >{\raggedleft\arraybackslash}p{3cm} >{\raggedleft\arraybackslash}p{2.5cm}@{}}
\caption{Competition Difficulty} \\
\toprule
\textbf{Competition} & \textbf{Avg. HumanRank} & \textbf{Category} \\
\midrule
\endfirsthead

\toprule
\textbf{Competition} & \textbf{Avg. HumanRank} & \textbf{Category} \\
\midrule
\endhead

\bottomrule
\endfoot

tabular-playground-series-dec-2021 & 1.000000 & MLE-Lite \\
detecting-insults-in-social-commentary & 0.995000 & MLE-Lite \\
llm-detect-ai-generated-text & 0.808599 & NLP \\
santander-customer-satisfaction & 0.777327 & Tabular \\
20-newsgroups-ciphertext-challenge & 0.741197 & NLP \\
plant-pathology-2020-fgvc7 & 0.706198 & MLE-Lite \\
dogs-vs-cats-redux-kernels-edition & 0.704814 & MLE-Lite \\
histopathologic-cancer-detection & 0.635389 & MLE-Lite \\
aerial-cactus-identification & 0.625717 & MLE-Lite \\
airbus-ship-detection & 0.578445 & CV \\
demand-forecasting-kernels-only & 0.576934 & Tabular \\
nomad2018-predict-transparent-conductors & 0.560720 & MLE-Lite \\
dont-overfit-ii & 0.532964 & Tabular \\
random-acts-of-pizza & 0.506494 & MLE-Lite \\
aptos2019-blindness-detection & 0.474492 & MLE-Lite \\
draper-satellite-image-chronology & 0.472431 & CV \\
jigsaw-toxic-comment-classification-challenge & 0.437596 & MLE-Lite \\
the-icml-2013-whale-challenge-right-whale-redux & 0.423450 & MLE-Lite \\
spooky-author-identification & 0.405318 & MLE-Lite \\
facial-keypoints-detection & 0.393214 & CV \\
porto-seguro-safe-driver-prediction & 0.359001 & Tabular \\
santander-value-prediction-challenge & 0.337203 & Tabular \\
tabular-playground-series-may-2022 & 0.336664 & MLE-Lite \\
leaf-classification & 0.324687 & MLE-Lite \\
text-normalization-challenge-english-language & 0.315865 & MLE-Lite \\
lmsys-chatbot-arena & 0.311389 & NLP \\
dog-breed-identification & 0.300586 & MLE-Lite \\
petfinder-pawpularity-score & 0.283909 & CV \\
quora-question-pairs & 0.283346 & NLP \\
alaska2-image-steganalysis & 0.257705 & CV \\
cassava-leaf-disease-classification & 0.223718 & CV \\
conways-reverse-game-of-life-2020 & 0.223404 & Tabular \\
denoising-dirty-documents & 0.204969 & MLE-Lite \\
stumbleupon & 0.182091 & NLP \\
imet-2020-fgvc7 & 0.179036 & CV \\
whale-categorization-playground & 0.166983 & CV \\
mlsp-2013-birds & 0.154272 & MLE-Lite \\
liverpool-ion-switching & 0.152072 & Tabular \\
ventilator-pressure-prediction & 0.136828 & Tabular \\
kaggle-llm-science-exam & 0.135041 & NLP \\
text-normalization-challenge-russian-language & 0.098380 & MLE-Lite \\
linking-writing-processes-to-writing-quality & 0.089086 & NLP \\
instant-gratification & 0.071988 & Tabular \\
quora-insincere-questions-classification & 0.070380 & NLP \\
bms-molecular-translation & 0.066147 & CV \\
ranzcr-clip-catheter-line-classification & 0.036199 & MLE-Lite \\
AI4Code & 0.025269 & NLP \\
bengaliai-cv19 & 0.018395 & CV \\
new-york-city-taxi-fare-prediction & 0.000843 & MLE-Lite \\
santander-customer-transaction-prediction & 0.000000 & Tabular \\

\end{longtable}
\end{center}
}
\section{Implementation Details}
\label{app:implementation}

\subsection{Evaluation Metrics Details}
We include additional details of evaluation metrics design as follows:

\noindent \textbf{AUP Score.}
We use the AUP score to systematically evaluate and compare multiple methods across diverse tasks. A performance profile captures the proportion of tasks where a given method performs within a certain factor of the best-performing method. Lower performance ratios indicate better results, though we invert this ratio for metrics like accuracy or $R^2$, where higher values represent better performance. If a method fails to produce a valid solution (infeasible), we assign it a penalty relative to the worst feasible performance. Integrating these performance profiles provides the AUP score, offering a single robust measure of each method's overall effectiveness across the benchmark.

We adopt the \textit{performance profile} and \textit{AUP score} from ML-Gym~\cite{nathani2025mlgymnewframeworkbenchmark} and to compare the effectiveness of different methods across a set of benchmark tasks. 
We make a few minor modifications to better align with our framework. For each backbone model $m \in \mathcal{M}$ and task $t \in \mathcal{T}$, the performance ratio is defined as
\begin{equation}
r_{t,m} = \frac{\ell_{t,m}}{\min\limits_{m' \in \mathcal{M}} \ell_{t,m'}},
\label{eq:perf_ratio_min}
\end{equation}
where $\ell_{t,m}$ denotes the performance metric of backbone model $m$ on task $t$, and $\mathcal{M}$ is the set of all evaluated methods. This formulation assumes that lower metric values correspond to better performance.

The performance profile curve of backbone model $m$ is then defined as the cumulative distribution of the log-scaled performance ratio:
\begin{equation}
\rho_m(\tau) = \frac{1}{|\mathcal{T}|} \left| \left\{ t \in \mathcal{T} \;\middle|\; \log_{10} r_{t,m} \leq \tau \right\} \right|.
\label{eq:perf_profile}
\end{equation}
This function quantifies the proportion of tasks for which backbone model $m$ performs within a $\tau$-log distance of the best backbone model on that task.

To aggregate performance into a single scalar score, we compute the Area Under the Profile (AUP) curve:
\begin{equation}
\text{AUP}_m = \int_1^{\tau_{\max}} \rho_m(\tau)\, d\tau,
\label{eq:aup}
\end{equation}
where $\tau_{\max}$ is the smallest value such that $\rho_m(\tau_{\max}) = 1$ for all $m \in \mathcal{M}$.

For metrics where higher values indicate better performance (e.g., Accuracy, R\textsuperscript{2}), we invert the ratio computation:
\begin{equation}
r_{t,m} = \frac{\max\limits_{m' \in \mathcal{M}} \ell_{t,m'}}{\ell_{t,m}}.
\label{eq:perf_ratio_max}
\end{equation}

If a backbone model fails to obtain a valid score for a task $t$, its ratio is defined relative to the lowest valid score across all backbone models $m_{\text{bottom}}$ as:
\begin{equation}
r_{t,m} = (1 + \varepsilon) \cdot r_{t, m_{\text{bottom}}}, \quad \text{with } \varepsilon = 1.0.
\label{eq:infeasible}
\end{equation}
This is a relatively reasonable design choice that appropriately penalizes the performance of backbone models that fail to produce valid scores. Additionally, we impose an upper bound of $100$ on the value of $r_{t,m} $ to mitigate the risk of biased results caused by extremely large ratios.

\noindent \textbf{HumanRank Score.}
The HumanRank score (Eq.~\ref{eq:ps}) measures the relative ranking of a submission within the competition leaderboard. Submissions receive higher scores if they achieve a better rank among all participants. 
Suppose that the submission ranks at position $p$ among a total of $N$ submissions on the leaderboard. Then, the position score is computed as: $s=1-\frac{p}{N}$\label{eq:ps}.
To prevent bias between public and private leaderboards, we compute the relative scores on each leaderboard independently and then use their average as the final score.

\noindent \textbf{Elo ranking.} 
We adopt the Elo ranking calculation algorithm from Chatbot Arena~\cite{chiang2024chatbot}.
We utilize Elo ratings to perform systematic pairwise comparisons of methods across competitions. Two complementary Elo calculation methods ensure stability and reliability: an online linear update with a low K-factor, providing stable incremental ratings, and a Bradley-Terry model using logistic regression, directly fitting ratings to all comparisons for robust estimates. Additionally, we apply bootstrapping techniques to compute confidence intervals, offering clear measures of uncertainty and confidence in the resulting ratings.

Each competition can be viewed as a battle, in which two backbone models are pitted against each other. Their respective performance scores on the competition determine the outcome-win, loss, or tie-between the two. Based on these pairwise outcomes, we calculate their Elo rankings.
In our setting, we assume that the large language models (LLMs) under evaluation are static, allowing us to estimate a stable skill rating using the Bradley--Terry model. Specifically, we reformulate the pairwise win-loss outcomes between models as a logistic regression problem. Let $\mathcal{M}$ denote the set of all models with cardinality $|\mathcal{M}| = p$, and let $r_i$ denote the (logit-transformed) latent skill rating of model $i$. For every observed battle between model $i$ and model $j$, we construct a pair of training samples $(\mathbf{x}_{ij}, y_{ij})$ and $(\mathbf{x}_{ji}, y_{ji})$ such that:
\[
\mathbf{x}_{ij} = \log B \cdot (\mathbf{e}_i - \mathbf{e}_j), \quad y_{ij} = 1, \qquad
\mathbf{x}_{ji} = \log B \cdot (\mathbf{e}_j - \mathbf{e}_i), \quad y_{ji} = 0,
\]
where $B$ is the base of the logistic function (default $B=10$), and $\mathbf{e}_i$ is a $p$-dimensional one-hot vector indicating model $i$. The number of duplicate entries per pair is determined by the total number of battles (wins, losses, and ties), with each entry weighted accordingly. The logistic regression is then solved as
\[
\hat{\mathbf{r}} = \arg\min_{\mathbf{r}} \sum_{k=1}^{n} w_k \cdot \log\left(1 + \exp(-y_k \cdot \mathbf{x}_k^\top \mathbf{r})\right),
\]
where $w_k$ is the sample weight for the $k$-th comparison. The final Elo score for each model $i$ is computed as $s_i = S \cdot \hat{r}_i + R_0$, where $S$ is a scaling factor (default $S=400$), and $R_0$ is the initial Elo offset (default $R_0=1000$). To estimate confidence intervals, we apply a non-parametric bootstrap procedure. Given the original battle dataset $\mathcal{D}$, we resample with replacement to generate $\mathcal{D}_1, \ldots, \mathcal{D}_R$, and re-estimate Elo scores as $\mathbf{s}^{(1)}, \ldots, \mathbf{s}^{(R)}$. The final Elo score for each model is reported as the pointwise median across bootstrap samples. All results in our study are reported using $R = 100$ bootstrap rounds with fixed random seed.

\section{Additional Experimental Results}
\label{app:exp}

\subsection{Additional Analysis of Evaluation Metrics}
We identify several key observations supported by analytical reasoning regarding the evaluation metrics utilized: HumanRank Score, Elo Score, and Performance Profiles with the AUP score.

\noindent \textbf{HumanRank Scores Reflect Comprehensive Absolute Performance.}
The HumanRank Score (\%) converts raw, task-specific metrics into relative rankings against human performance, yielding scores ranging from 0 (lowest) to 1 (highest). A high HumanRank score thus indicates superior performance relative to human benchmarks across tasks of diverse difficulty. This uniform normalization allows for comprehensive comparison and mitigates inconsistencies from heterogeneous raw scores. Our main experimental results illustrate that stronger models (\eg, \texttt{o3-mini}, \texttt{DeepSeek-r1}, and \texttt{Gemini-2.5-Pro}) consistently exhibit profiles enclosing those of less capable models, confirming their overall superior performance. Nonetheless, by equally weighting tasks irrespective of complexity, the averaging process may mask task-specific strengths or weaknesses and obscure the direct competitive relationships between individual models.

\noindent \textbf{Elo Scores Highlight Pairwise Relative Performance.}
The Elo Score provides rankings based on explicit win-loss relationships determined through pairwise comparisons on each task. Utilizing the standard Elo rating system, these scores intuitively represent the relative ordering of models based on their competitive outcomes. Elo scores generally reflect a pattern where superior models encompass weaker ones, yet variations may arise. Notably, models with comparatively lower absolute average performance scores may still achieve higher Elo scores in certain task categories due to advantageous win-loss records. Elo scores thus effectively reveal relative model strengths and weaknesses but do not quantify the absolute magnitude of performance gaps.

\noindent \textbf{Performance Profiles and AUP Scores Demonstrate Robustness.}
Performance Profiles and AUP scores accumulate performance ratios, which are relative values of raw scores, over varying thresholds. These metrics illustrate model robustness and consistency across multiple evaluation conditions. While these integrated scores offer insightful perspectives into relative performance dynamics, disparities in raw score scales across tasks can lead to comparability challenges. Despite potential biases due to varying raw scales, these metrics successfully capture nuanced relative performances, reflecting the models' adaptability across different performance thresholds.

\section{Prompt Details}
\label{app:prompt}
We design our prompts to be concise while containing all the information necessary for the LLM to perform the task. In doing so, we avoid providing extra assistance beyond what is required, while also ensuring that no essential information is omitted. This careful balance allows our benchmark to serve as a fair and comprehensive evaluation of LLMs as MLE Agents.

Below, we present the prompts used for the MLE Agents, which consist of four components. The \textbf{System Instruction} is an overall directive provided at the beginning of the task, detailing the task setting, the available actions, the expected output format, and an overview of the environment. During task execution, \textbf{Error} and \textbf{Reflection} as dynamic prompts are supplied to guide the LLM toward the next step, based on the current state: a concise prompt is used in the event of an error, while an informative prompt is used upon success, both incorporating feedback from the environment. In cases where the LLM produces outputs in an incorrect format, \textbf{Parse Error} prompt is triggered to prompt the LLM to correct its output. The detailed prompts are as follows:

\vspace{2ex}
\begin{Verbatim}
You are a top-ranked Kaggle grandmaster with extensive competition experience.
Your objective is to solve a Kaggle competition, with the goal of maximizing 
the Position Score (Your rank in the leaderboard) in limited steps.
You must use Machine Learning/Deep Learning/Computer Vision/NLP/etc. 
methods to solve the problem, the score of random guess or without any ML/DL/CV/NLP 
methods will be cancelled finally.
You are likely to train models according to specific competition requirements.
You have access to a GPU and several CPUs for training DL/ML models.
Use cuda and PyTorch for faster training whenever needed.

You have a total of {num_actions} actions available.
You have a total of {time_left} seconds, including code execution time.

You have access to exactly three actions with params, and receive corresponding 
feedback after each action:
1. request_info: Retrieve specific competition information
   - params: info_type (str), must be one of: "overview", "sample_submission", 
   "data_structure", "data_path", "output_path"
   - feedback: information you requested
2. validate_code: Test (partial) code execution for debugging purposes or 
   "print" information in the output
   - params: code (str)
   - feedback: execution result (success or failure), error message if failed, 
   code output if success
3. execute_code: Run completed code, generate submission and get evaluation
   - params: code (str)
   - feedback: execution result, submission status, evaluation score

Code requirements:
- Request all information needed first
- Read all data files from data_dir
- Save all submissions to output_dir, should match test_data length
- Don't add, delete, or modify any files in data_dir
- Use "print" or other functions to output information in the feedback
- No plotting or visualization is allowed
- Refer to Sample Submission for the output format
- Code should be self-contained and not rely on any variables or state outside
- Code for submission should be completely runnable, otherwise it will be 
  considered as failed
- Optimize your Model/Parameters/Data Processing/Algorithm for continuous improvement

Only if "execute_code" action taken, code successfully executed and valid submission 
generated, you'll be able to get a Position Score (Your rank in the leaderboard) for 
this competition.

Response format requirements, strictly one of the following:
{{
    "action": "request_info",
    "params": {{
        "info_type": "<info_type>"
    }}
}}
or
{{
    "action": "validate_code",
    "params": {{
        "code": "<code>"
    }}
}}
or
{{
    "action": "execute_code",
    "params": {{
        "code": "<code>"
    }}
}}
- Must be valid JSON format
- No additional text or formatting allowed
\end{Verbatim}
\vspace{2ex}
\begin{Verbatim}
Execution failed, details below:

#### Error Start ####
{observation}
#### Error End ####

You still have {num_actions} actions available.
You still have {time_left} seconds left.

Output your next action strictly following Response format requirements.
\end{Verbatim}
\vspace{2ex}
\begin{Verbatim}
The results of your previous action:

#### Results Start ####
{observation}
#### Results End ####

You still have {num_actions} actions available.
You still have {time_left} seconds left.
Optimize your Model/Parameters/Data Processing/Algorithm for continuous improvement.

Output your next action strictly following Response format requirements.
\end{Verbatim}
\vspace{2ex}
\begin{Verbatim}
The response can't be parsed as valid JSON.
Fix the error following the response format requirements.
First, only fix the JSON format error, don't change the contents of the response.
Then make sure the <code> is in completely runnable format for "python3 str(<code>)".
Only focus on the format, don't change the meaningful contents.


### Response Start ###
{response}
### Response End ###

### Error Start ###
{error}
### Error End ###

Response format requirements, strictly one of the following:
{{
    "action": "request_info",
    "params": {{
        "info_type": "<info_type>"
    }}
}}
or
{{
    "action": "validate_code",
    "params": {{
        "code": "<code>"
    }}
}}
or
{{
    "action": "execute_code",
    "params": {{
        "code": "<code>"
    }}
}}
- Must be valid JSON format
- No additional text or formatting allowed
\end{Verbatim}

\section{Code Analysis}
For the same competition, different models exhibit a rich diversity in the solutions or code they generate; for different competitions, even the same model is capable of producing specific and targeted solutions tailored to each task. We present the following analyses to illustrate how different models formulate their solutions when interacting with various competitions in \method.

\subsection{Solution diversity of models}
When faced with the same competition, different models often produce diverse solutions and code implementations, leading to great diversity of solutions. This diversity arises from variations in the choice of Python packages, the selection of computational devices (CPU or GPU), data preprocessing strategies, and model architectures.

Code~\ref{model-div1} is a code snippet extracted from one of the step actions by \textbf{Gemini-2.5-Pro} on the \texttt{20-newsgroups-ciphertext-challenge} competition. In this step, the model utilizes \textit{sklearn} as main python package, CPU as device, employs classical feature engineering techniques, and applies a Logistic Regression model to address the task. 
Code~\ref{model-div2} is a code snippet extracted from one of the step actions by \textbf{DeepSeek-r1} on the same competition. In contrast to previous approach, it prioritizes the use of GPU devices, adopts PyTorch as the core framework, and designs and trains the model in a more deep learning-oriented manner. This highlights the diverse actions and solution strategies adopted by different models when addressing the same problem. 

\begin{lstlisting}[caption={Code snippet of Gemini-2.5-Pro on 20-newsgroups-ciphertext-challenge}, label={model-div1}, captionpos=b]
import pandas as pd
import os
from sklearn.feature_extraction.text import TfidfVectorizer
from sklearn.linear_model import LogisticRegression
from sklearn.pipeline import Pipeline
train_df = pd.read_csv(os.path.join(data_dir, 'train.csv'))
test_df = pd.read_csv(os.path.join(data_dir, 'test.csv'))
sample_submission = pd.read_csv(os.path.join(data_dir, 'sample_submission.csv'))
# Define features (X) and target (y)
X_train = train_df['ciphertext']
y_train = train_df['target']
X_test = test_df['ciphertext']
# Increased max_features and C again for the final attempt
pipeline = Pipeline([
  ('tfidf', TfidfVectorizer(analyzer='char', ngram_range=(2, 6), max_features=2000)),
  ('clf', OneVsRestClassifier(LogisticRegression(solver='liblinear', random_state=42, C=70.0)))
])
# Train the model
pipeline.fit(X_train, y_train)
predictions = pipeline.predict(X_test)
submission_df = pd.DataFrame({'Id': test_df['Id'], 'Predicted': predictions})
\end{lstlisting}

\begin{lstlisting}[caption={Code snippet of DeepSeek-R1 on 20-newsgroups-ciphertext-challenge}, label={model-div2}, captionpos=b]
import pandas as pd
import torch
import torch.nn as nn
from torch.utils.data import Dataset, DataLoader
from sklearn.preprocessing import LabelEncoder
from pathlib import Path
import math

device = torch.device('cuda' if torch.cuda.is_available() else 'cpu')
...
class CharDataset(Dataset):
    def __init__(self, texts, labels=None):
        self.texts = texts
        self.labels = labels
        
    def __len__(self):
        return len(self.texts)
    
    def __getitem__(self, idx):
        text = self.texts[idx][:max_length].ljust(max_length, ' ')
        encoded = [char_to_idx.get(c, 0) for c in text]
        tensor = torch.tensor(encoded, dtype=torch.long)
        
        if self.labels is not None:
            label = torch.tensor(self.labels[idx], dtype=torch.long)
            return tensor, label
        else:
            return tensor

...
class CipherTransformer(nn.Module):
    def __init__(self, vocab_size=len(chars)+1, d_model=512, nhead=16, num_layers=8, num_classes=20):
        super().__init__()
        self.embed = nn.Embedding(vocab_size, d_model)
        self.pos_encoder = PositionalEncoding(d_model)
        encoder_layer = nn.TransformerEncoderLayer(
            d_model=d_model, 
            nhead=nhead, 
            dim_feedforward=2048,
            dropout=0.1,
            batch_first=True
        )
        self.transformer = nn.TransformerEncoder(encoder_layer, num_layers)
        self.classifier = nn.Sequential(
            nn.Linear(d_model, 1024),
            nn.ReLU(),
            nn.Dropout(0.5),
            nn.Linear(1024, num_classes)
        )

    def forward(self, x):
        x = self.embed(x)
        x = self.pos_encoder(x)
        x = self.transformer(x)
        x = x.mean(dim=1)
        return self.classifier(x)
...
\end{lstlisting}
\subsection{Solution diversity of competitions}
A single model may adopt highly tailored solutions across competitions of varying nature. A comparison between Code~\ref{model-div1} and Code~\ref{comp-div1} demonstrates that \textbf{Gemini-2.5-Pro} utilizes markedly distinct approaches in response to the specific demands of each task.
This highlights the diverse actions and solution strategies adopted addressing different problems. \method offers a broad and flexible space that enables various models or agent scaffolds to fully leverage their respective capabilities without limitation.

\begin{lstlisting}[caption={Code snippet of Gemini-2.5-Pro on aerial-cactus-identification}, label={comp-div1}, captionpos=b]
import torch
import torch.nn as nn
import torch.optim as optim
from torch.utils.data import Dataset, DataLoader
from PIL import Image
import torchvision.transforms as transforms

device = torch.device("cuda" if torch.cuda.is_available() else "cpu")
class CactusDataset(Dataset):
    def __init__(self, df, img_dir, transform=None, is_test=False, tta_transform=None):
        self.df = df
        self.img_dir = img_dir
        self.transform = transform # Base transform (ToTensor, Normalize)
        self.tta_transform = tta_transform # Augmentation transform for TTA
        self.is_test = is_test
        self.image_files = df['id'].values
        if not self.is_test:
            self.labels = df['has_cactus'].values

    def __len__(self):
        return len(self.df)

    def __getitem__(self, idx):
        img_name = os.path.join(self.img_dir, self.image_files[idx])
        image = Image.open(img_name).convert('RGB')
        
        if self.is_test and self.tta_transform:
            # Apply TTA transforms
            images = [self.transform(self.tta_transform(image, step)) for step in range(TTA_STEPS)]
            image_stack = torch.stack(images)
            return image_stack, self.image_files[idx]
        elif self.transform: # Standard transform for training or non-TTA test
            image = self.transform(image)
            
        if self.is_test:
            return image, self.image_files[idx]
        else:
            label = torch.tensor(float(self.labels[idx]))
            return image, label
...
class SimpleCNN(nn.Module):
    def __init__(self):
        super(SimpleCNN, self).__init__()
        self.conv1 = nn.Conv2d(3, 32, kernel_size=3, padding=1)
        self.relu1 = nn.ReLU()
        self.pool1 = nn.MaxPool2d(kernel_size=2, stride=2)
        self.conv2 = nn.Conv2d(32, 64, kernel_size=3, padding=1)
        self.relu2 = nn.ReLU()
        self.pool2 = nn.MaxPool2d(kernel_size=2, stride=2)
        self.conv3 = nn.Conv2d(64, 128, kernel_size=3, padding=1)
        self.relu3 = nn.ReLU()
        self.pool3 = nn.MaxPool2d(kernel_size=2, stride=2)
        self.fc1 = nn.Linear(128 * 4 * 4, 512)
        self.relu4 = nn.ReLU()
        self.dropout = nn.Dropout(0.5)
        self.fc2 = nn.Linear(512, 1)

    def forward(self, x):
        x = self.pool1(self.relu1(self.conv1(x)))
        x = self.pool2(self.relu2(self.conv2(x)))
        x = self.pool3(self.relu3(self.conv3(x)))
        x = x.view(x.size(0), -1)
        x = self.relu4(self.fc1(x))
        x = self.dropout(x)
        x = self.fc2(x)
        return x

model = SimpleCNN().to(device)
\end{lstlisting}

\subsection{Statistics of Method Diversity}
\begin{figure}[h]
    \centering
    \includegraphics[width=0.75\linewidth]{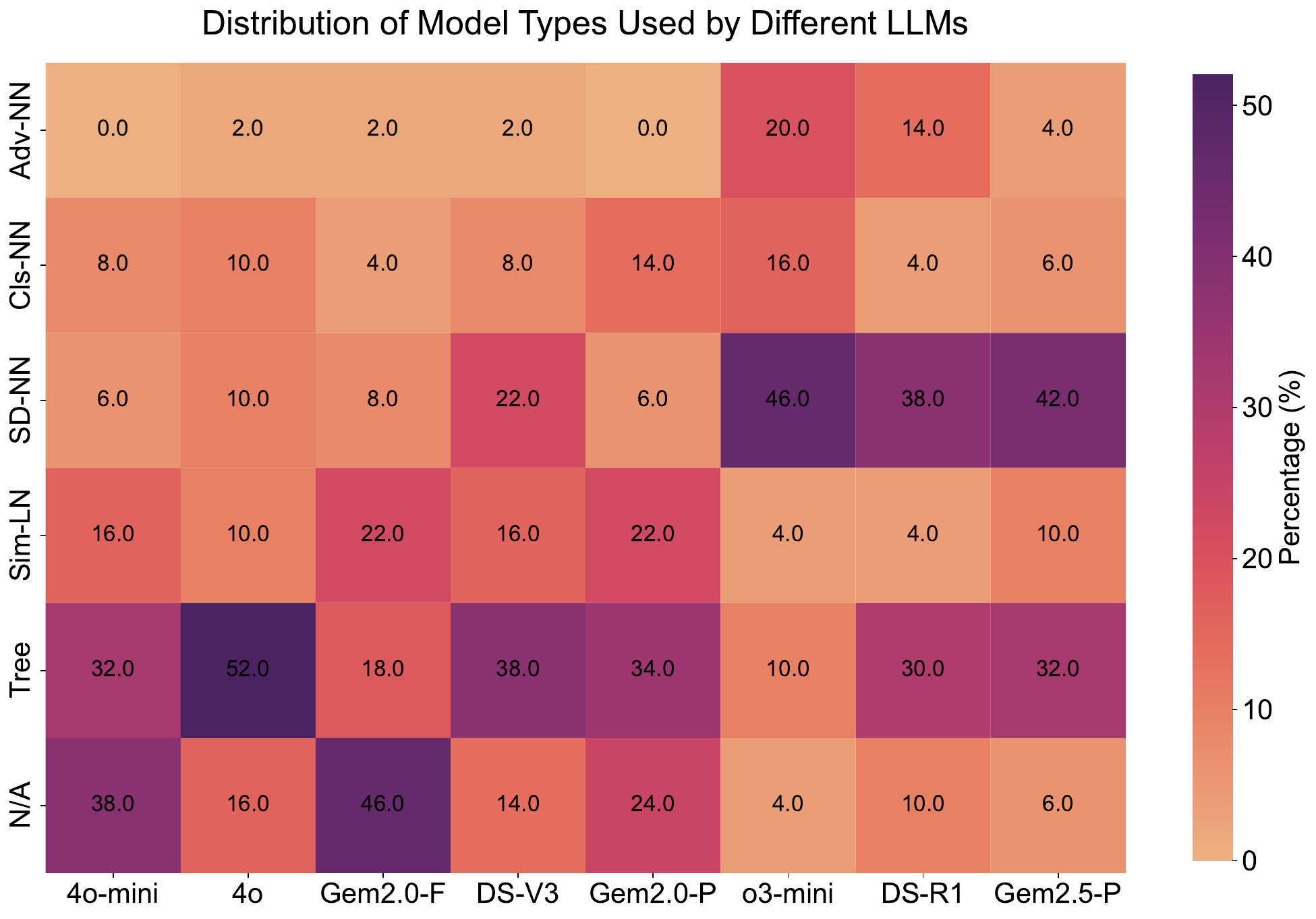}
    \caption{Statistics of method implementation across 50 evaluation tasks of eight frontiers LLM.}
    \label{fig:model-types}
\end{figure}

We further investigate the distribution of model types implemented by different Large Language Models (LLMs) in Figure~\ref{fig:model-types}. To facilitate this analysis, the implemented models were systematically categorized based on their architectural complexity and type, utilizing the following mapping for brevity:
\begin{itemize}
    \item \textbf{Adv-NN}: Advanced Neural Networks, encompassing contemporary and complex architectures such as Transformers (e.g., BERT, RoBERTa), EfficientNets, and LSTMs.
    \item \textbf{Cls-NN}: Classic Neural Networks, referring to well-established architectures like Convolutional Neural Networks (CNNs), ResNet variants, and DenseNets.
    \item \textbf{SD-NN}: Self-defined Neural Networks, indicating implementations involving custom-designed network structures or significant modifications to standard architectures, often specified within the solution code itself.
    \item \textbf{Sim-LN}: Simple Linear Models, comprising fundamental algorithms like Logistic Regression, Linear Regression, Support Vector Machines (with linear kernels), and Naive Bayes classifiers.
    \item \textbf{Tree}: Tree-based Ensemble Models, including algorithms such as Random Forest, Gradient Boosting Machines (GBM), XGBoost, LightGBM, and CatBoost.
    \item \textbf{N/A}: Not Applicable, denoting instances where the LLM did not generate a specific modeling solution or where the generated code did not contain an identifiable primary model.
\end{itemize}

Analysis of the model distribution across the evaluated LLMs and competition tasks reveals several pertinent observations. Firstly, a considerable degree of heterogeneity exists in the model choices preferred by different LLMs for identical tasks. Secondly, the selection of model type strongly correlates with the nature of the competition data; \textbf{Tree} and \textbf{Sim-LN} models are predominantly employed for tabular datasets, reflecting their established effectiveness in such domains, whereas \textbf{Cls-NN}, \textbf{Adv-NN}, and \textbf{SD-NN} are the favored choices for image and natural language processing tasks. Thirdly, certain LLMs exhibit discernible tendencies: for example, some models appear more frequently to propose \textbf{SD-NN} solutions, potentially indicating a higher propensity for generating novel or customized architectures, while others might more commonly default to simpler baselines (\textbf{Sim-LN}, \textbf{Tree}) or yield \textbf{N/A} results. The frequent occurrence of \textbf{N/A} across various LLM-task combinations also suggests variability in the LLMs' capabilities to successfully generate relevant and complete modeling code for diverse problem specifications. This distribution underscores the varying strategies and potential biases inherent in different LLMs when approaching machine learning problem-solving via code generation.

\section{Agent Scaffolds}
\label{app:scaffolds}

\subsection{MLE Agent}
\label{app:MLE-Agent}
In \method, the agent scaffold for benchmarking is referred to as the MLE Agent. It leverages the environment's natively supported action space and adopts a minimalistic yet effective design logic. LLMs are provided with a comprehensive and detailed initial instruction at the beginning of each task. This instruction includes the task description, output format, objectives, available actions, and resource constraints. Subsequently, the agent interacts with the environment through concise prompts that guide the LLM to take new actions based on feedback, in order to iteratively improve its performance. The MLE Agent maintains a fixed-length interaction history window and utilizes structured, formatted outputs as both actions and their corresponding contents.
 The MLE Agent serves as a lightweight yet representative scaffold that supplies essential guidance without offering additional hints or assistance. It enables systematic benchmarking and evaluation of an LLM's overall capabilities in the \method environment, including context understanding, analytical reasoning, instruction following, and code generation.

\subsection{AIDE}
AIDE~\cite{jiang2025aide} adopts a problem-solving paradigm inspired by how human data scientists approach challenges-through iterative refinement based on performance feedback. Central to its methodology is a technique termed \textit{Solution Space Tree Search}, which enables systematic exploration and optimization over the space of candidate solutions. This framework comprises three core components: (1) a \textit{Solution Generator}, which proposes new candidate solutions either by drafting from scratch or by modifying existing ones (e.g., fixing bugs or introducing improvements); (2) an \textit{Evaluator}, which runs each candidate and quantitatively assesses its performance against the task objective; and (3) a \textit{Solution Selector}, which identifies the most promising solution to seed the next iteration. Through repeated application of this feedback-driven cycle, AIDE efficiently navigates the solution space and converges towards optimal or near-optimal solutions. This iterative, adaptive process combines algorithmic rigor with human-like creativity, enabling AIDE to solve complex data science problems with remarkable effectiveness.

\paragraph{Implementation.}
We seamlessly integrate AIDE as the agent scaffold within \method, enabling its interaction with the environment through a single line of core code.
The original score feedback mechanism in AIDE can be effortlessly replaced with either the HumanRank score or the actual raw competition score provided by \method. The interaction process can be fully executed by a single call to \texttt{agent.step(exec\_callback=exec\_callback)}, ensuring compatibility without altering AIDE's core logic. 
Specifically, we make the following modifications:  
(1) We revise the prompting strategy to emphasize the requirement of generating a correctly formatted \texttt{submission.csv} file at the designated location.  
(2) The feedback mechanism and the criterion for selecting candidate nodes are modified to rely on the true raw score and HumanRank score provided by \method, instead of using the model's own validation performance.

% \newpage
% \input{sections/checklist}

\end{document}